\documentclass{article}
\usepackage{arxiv}
\usepackage[utf8]{inputenc} 
\usepackage[brazil]{babel}
\usepackage[T1]{fontenc}    
\usepackage{hyperref}       
\usepackage{url}            
\usepackage{booktabs}       
\usepackage[backend=biber, style=apa, sorting=nyt, language=english]{biblatex} 
\DeclareFieldFormat{journaltitle}{\textbf{#1}}

\usepackage{amsfonts}       
\usepackage{nicefrac}       
\usepackage{microtype}      
\usepackage{lipsum}
\usepackage{graphicx}
\usepackage{amsmath}
\usepackage{csquotes}
\usepackage{ragged2e} 
\usepackage{subcaption}
\usepackage{float}  
\usepackage[ruled,vlined,linesnumbered]{algorithm2e}
\graphicspath{ {./images/} }

\title{Redes Neurais com LSTM e GRU na Modelagem de Focos Ativos na Amazônia}

\author{
  Ramon Lima de Oliveira Tavares \\
  Departamento de Estatística\\
  Universidade Estadual da Paraíba\\
  \texttt{ramon.tavares@aluno.uepb.edu.br} \\
  \And
  Ricardo Alves de Olinda \\
  Departamento de Estatística\\
  Universidade Estadual da Paraíba\\
  \texttt{prof\_ricardo@servidor.uepb.edu.br} \\
}

\begin{document}
\maketitle
\begin{center}
	\large{\textbf{Resumo}}
\end{center}
\noindent
Este estudo apresenta uma metodologia abrangente para modelagem e previsão da série temporal histórica de focos ativos detectados pelo satélite \textit{AQUA\_M-T} na Amazônia, Brasil. A abordagem utiliza um modelo misto de Redes Neurais Recorrentes (RNN), combinando as arquiteturas \textit{Long Short-Term Memory} (LSTM) e \textit{Gated Recurrent Unit} (GRU) para prever os acumulados mensais de focos ativos detectados diariamente. A análise dos dados revelou uma sazonalidade consistente ao longo do tempo, com os valores máximos e mínimos anuais tendendo a se repetir nos mesmos períodos a cada ano. O objetivo principal é verificar se as previsões capturam essa sazonalidade inerente por meio de técnicas de aprendizado de máquina. A metodologia envolveu uma cuidadosa preparação dos dados, configuração do modelo e treinamento utilizando validação cruzada com duas sementes, garantindo que os dados se generalizem bem para os conjuntos de teste e validação para ambas as sementes. Os resultados apontam que o modelo combinado de LSTM e GRU oferece excelentes resultados nas previsões, demonstrando sua eficácia na captura de padrões temporais complexos e na modelagem da série temporal observada. Esta pesquisa contribui significativamente para a aplicação de técnicas de aprendizado profundo no monitoramento ambiental, especificamente na previsão de focos ativos. A abordagem proposta destaca o potencial de adaptação para outros desafios de previsão em séries temporais, abrindo novas oportunidades para pesquisa e desenvolvimento em aprendizado de máquina e previsão de fenômenos naturais.

\textbf{Palavras-chave:} Previsão de Séries Temporais; Redes Neurais Recorrentes; Aprendizado Profundo.
\\

\begin{abstract}
	This study presents a comprehensive methodology for modeling and forecasting the historical time series of active fire spots detected by the AQUA\_M-T satellite in the Amazon, Brazil. The approach employs a mixed Recurrent Neural Network (RNN) model, combining Long Short-Term Memory (LSTM) and Gated Recurrent Unit (GRU) architectures to predict the monthly accumulations of daily detected active fire spots. Data analysis revealed a consistent seasonality over time, with annual maximum and minimum values tending to repeat at the same periods each year. The primary objective is to verify whether the forecasts capture this inherent seasonality through machine learning techniques. The methodology involved careful data preparation, model configuration, and training using cross-validation with two seeds, ensuring that the data generalizes well to both the test and validation sets for both seeds. The results indicate that the combined LSTM and GRU model delivers excellent forecasting performance, demonstrating its effectiveness in capturing complex temporal patterns and modeling the observed time series. This research significantly contributes to the application of deep learning techniques in environmental monitoring, specifically in forecasting active fire spots. The proposed approach highlights the potential for adaptation to other time series forecasting challenges, opening new opportunities for research and development in machine learning and prediction of natural phenomena.
	
\end{abstract}


\section{Introdução}
Séries temporais são amplamente utilizadas em diversas áreas, como economia, climatologia, e monitoramento ambiental, e contam com grandes referências como \cite{BoxJenkins1976}, \cite{JamesHamilton1994}, e \cite{PeterBrockwell2002}. De maneira geral, uma série temporal pode ser definida como um conjunto de informações fixadas no tempo e/ou no espaço de forma padronizada ou não. Quando tratamos de séries temporais de dados quantitativos discretos, onde o tempo é o principal fator de interesse, podemos entender essa série como um conjunto de observações que representam quantidades específicas, registradas ao longo do tempo. No contexto deste trabalho, focamos na série temporal dos focos ativos detectados pelo satélite \textit{AQUA\_M-T} na Amazônia, Brasil. Os focos ativos são detectados com base em anomalias de temperatura em pixels observados pelo satélite. Quando a temperatura de um pixel (representa uma área de 1m²) atinge níveis significativamente elevados, como, por exemplo, acima de 47°C — valor que, segundo o Sistema Estadual de Informações Ambientais e Recursos Hídricos \cite{SEIA2024}, caracteriza um foco de calor — o satélite registra a ocorrência de um foco ativo. Esses dados, disponibilizados mensalmente pelo Instituto Nacional de Pesquisas Espaciais (INPE), oferecem uma visão histórica importante, embora seja sabido que os satélites como o \textit{AQUA\_M-T} possuem limitações em termos de precisão, devido à sua idade e tecnologia. Entretanto, mesmo com essas limitações, os dados são valiosos para a identificação de padrões sazonais e anomalias ao longo dos anos. Futuramente, espera-se que esses dados sejam atualizados com a entrada em operação de novos satélites, o que permitirá um monitoramento ainda mais preciso. Neste trabalho, utilizamos modelos de Redes Neurais Recorrentes (RNNs), especificamente as arquiteturas \textit{Long Short-Term Memory} (LSTM) propostas por \cite{Schmidhuber1997} e \textit{Gated Recurrent Unit} (GRU) propostas por \cite{JunyoungChung2014}, para modelar e prever a quantidade de focos ativos na Amazônia. A LSTM é conhecida por sua capacidade de lidar com problemas de retenção de longo prazo, enquanto a GRU simplifica a estrutura da LSTM, aumentando a eficiência do modelo. A combinação dessas duas arquiteturas em um modelo misto oferece a robustez necessária para capturar os padrões complexos presentes na série temporal analisada. O lado positivo das redes neurais é a capacidade de um modelo bem treinado aprender padrões independentemente da escala dos dados. No caso dos focos ativos, a série temporal varia de um mínimo de 70 focos registrados em abril de 1999 a um máximo de 73.141 focos em setembro de 2007. Esse intervalo expressivo demonstra a importância de desenvolver uma arquitetura robusta e bem configurada para garantir que o modelo consiga aprender essas variações e realizar previsões com menores erros em comparação aos valores reais observados. Neste artigo, além de explorar a aplicação das RNNs, LSTM e GRU, você encontrará uma visão detalhada de como foi estruturado e treinado o modelo, quantos neurônios e épocas de treinamento foram utilizados, e como as previsões foram realizadas. Discutiremos a fundamentação teórica por trás das redes neurais recorrentes, analisaremos os dados históricos, identificando a sazonalidade dos focos ativos na Amazônia, e apresentaremos os resultados das previsões geradas pelo modelo treinado. Por fim, serão discutidas as implicações dessas previsões e as conclusões deste estudo. Dito isso, seguimos adiante com este estudo, detalhando cada etapa do processo de modelagem, treinamento, validação e previsão, para demonstrar a eficácia das redes neurais recorrentes na análise de séries temporais ambientais.

\section{Referencial Teórico}

De acordo com \cite{Graves2013}, as Redes Neurais Recorrentes \textit{Recurrent Neural Networks} (RNNs) são modelos poderosos para dados sequenciais. Elas são capazes de lidar com problemas de rotulagem de sequências onde o alinhamento entre entrada e saída é desconhecido. Esses modelos são construídos para aprender dependências temporais em dados sequenciais e mantêm uma memória interna para processar informações anteriores.

\subsection{Unidade RNN}
Dada uma sequência de entrada \((x_{t}, x_{t+1}, x_{t+2}, \ldots, x_{t+n})\), uma Rede Neural Recorrente padrão computa a sequência de vetores ocultos \((h_{t-1}, h_{t}, h_{t+1}, h_{t+2}, \ldots, h_{t+n})\) e a sequência de vetores de saída \((y_{t}, y_{t+1}, y_{t+2}, \ldots, y_{t+n})\). 

\begin{equation}\label{eq:rnn1}
\mathbf{h}_t = \mathbf{A}(\mathbf{W}_{hh} \odot \mathbf{h}_{t-1} + \mathbf{W}_{xh} \odot \mathbf{x}_t + \mathbf{b}_h),
\end{equation}
\begin{equation}\label{eq:rnn2}
\mathbf{y}_t = (\mathbf{W}_{yi} \odot \mathbf{h}_{t} + \mathbf{b}_{y}),
\end{equation}

em que \textbf{W} é a matriz de pesos e \textbf{b} é o viés, e o operador $\odot$ representa a multiplicação elemento a elemento; o estado de saída $y_t$ gerado no tempo $t$ é determinado pela informação de entrada $x_t$ e pelo estado oculto anterior $h_{t-1}$ no tempo $t-1$. 

\begin{figure}[H] 
\centering
\caption{Esquema detalhado da unidade de Rede Neural Recorrente (RNN) simples}
\includegraphics[width=0.9\linewidth]{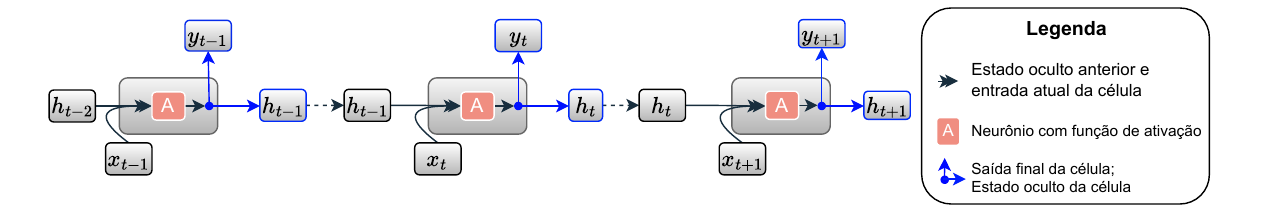}
\parbox{\linewidth}{\centering\noindent{Fonte: Elaborado pelo autor, adaptado de \cite{Greff2017}.}}
\label{fig:rnn}
\end{figure}

A Equação \eqref{eq:rnn1} mostra como o estado oculto atual $h_t$ é calculado usando uma função de ativação $\mathbf{A}$, pesos $\mathbf{W}$ e viés $\mathbf{b}$ correspondentes. Esse modelo de unidade de Redes Neurais Recorrentes é fundamental para compreender a propagação de informações ao longo do tempo em uma Rede Neural Recorrente. A estrutura interna da unidade RNN é exibida na Figura \ref{fig:rnn}.

\subsection{Unidade LSTM}

No artigo \textit{Speech Recognition with Deep Recurrent Neural Networks} \cite{Graves2013}, os autores enfatizam que a arquitetura das Redes de Memória de Longo e Curto Prazo (\textit{Long Short-Term Memory}, LSTM) é particularmente eficaz para tarefas que requerem o processamento de sequências temporais longas. As LSTMs se destacam pela capacidade de superar as limitações das Redes Neurais Recorrentes (RNNs) tradicionais, permitindo que informações relevantes sejam retidas por períodos mais prolongados. Isso é primordial para lidar com dependências temporais extensas. Enquanto as RNNs funcionam em sequências temporais mantendo uma memória interna, as LSTMs aprimoram essa capacidade ao utilizar \textit{gates} \enquote{portões} para controlar o fluxo de informações. Esses portões facilitam uma retenção mais eficaz das informações a longo prazo, comparado às RNNs tradicionais, que enfrentam dificuldades em manter dependências temporais mais longas. Dessa forma, as LSTMs demonstram uma capacidade superior de generalização e previsão quando confrontadas com dados de entrada que se estendem por longos períodos de tempo.

\begin{figure}[H] 
\centering
\caption{Esquema detalhado da unidade LSTM}
\includegraphics[width=0.8\linewidth]{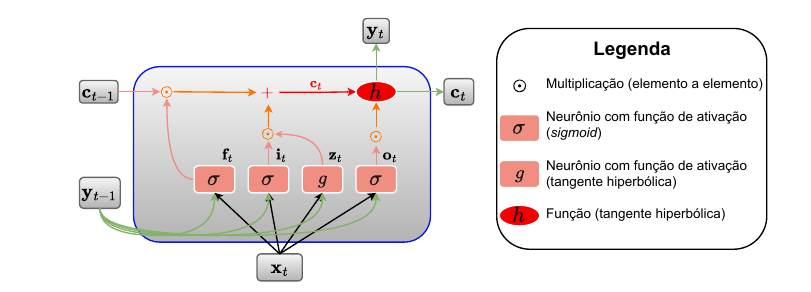}
\parbox{\linewidth}{\centering\noindent{Fonte: Elaborado pelo autor, adaptado de \cite{Greff2017}.}}
\label{fig:lstm}
\end{figure}

A arquitetura \textit{Long Short-Term Memory} (LSTM), conforme descrito por \cite{Greff2017}, é projetada para lidar com as limitações das Redes Neurais Recorrentes tradicionais em tarefas de aprendizado de sequências temporais. O bloco LSTM é composto por três componentes principais, como ilustrado na Figura \ref{fig:lstm}:

\begin{itemize}
    \item \textbf{Portão de Entrada:} Este portão regula a quantidade de nova informação que será incorporada na célula de memória. Ele determina quais informações devem ser adicionadas ao estado da célula.
    \item \textbf{Portão de Esquecimento:} Este portão decide quais informações presentes na célula de memória devem ser descartadas. Ele ajuda a manter a relevância dos dados ao longo do tempo, removendo informações que não são mais necessárias.
    \item \textbf{Portão de Saída:} Este portão controla a quantidade de informação da célula de memória que será utilizada na saída do bloco LSTM. Ele decide quais informações da célula de memória serão passadas para a próxima etapa na sequência.
\end{itemize}

Esses portões são responsáveis por regular o fluxo de informações dentro do bloco LSTM, permitindo a retenção e atualização eficaz de dados relevantes por longos períodos. A estrutura interna do LSTM permite que o modelo capture dependências temporais extensas e mantenha a precisão em tarefas que envolvem sequências longas e complexas.

Seja $\mathbf{x}_t$ o vetor de entrada no tempo $t$, $N$ o número de unidades LSTM na camada e $M$ o número de entradas (aqui $N \times M$ representa a dimensão da matriz de pesos). Então, obtemos os seguintes pesos para uma camada LSTM:

\begin{itemize}
\item Pesos de entrada: $\mathbf{W}_z, \mathbf{W}_i, \mathbf{W}_f, \mathbf{W}_o \in \mathbb{R}^{N\times M}$;
\item Pesos recorrentes: $\mathbf{R}_z, \mathbf{R}_i, \mathbf{R}_f, \mathbf{R}_o \in \mathbb{R}^{N\times N}$;
\item Pesos de viés: $\mathbf{b}_z, \mathbf{b}_i, \mathbf{b}_f, \mathbf{b}_o \in \mathbb{R}^{N}$.
\end{itemize}

Então, de acordo com \cite{Greff2017}, as fórmulas vetoriais para uma passagem direta em uma camada LSTM podem ser escritas como:

\begin{equation}
\bar{\mathbf{z}}_t = \mathbf{W}_z \mathbf{x}_t + \mathbf{R}_z \mathbf{y}_{t-1} + \mathbf{b}_z,
\end{equation}
\begin{equation*}
\mathbf{z}_t = g(\bar{\mathbf{z}}_t) \quad \text{\textit{entrada do bloco}};
\end{equation*}
\begin{equation}
\bar{\mathbf{i}}_t = \mathbf{W}_i \mathbf{x}_t + \mathbf{R}_i \mathbf{y}_{t-1} + \mathbf{b}_i,
\end{equation}
\begin{equation*}
\mathbf{i}_t = \sigma(\bar{\mathbf{i}}_t) \quad \text{\textit{porta de entrada}};
\end{equation*}
\begin{equation}
\bar{\mathbf{f}}_t = \mathbf{W}_f \mathbf{x}_t + \mathbf{R}_f \mathbf{y}_{t-1} + \mathbf{b}_f,
\end{equation}
\begin{equation*}
\mathbf{f}_t = \sigma(\bar{\mathbf{f}}_t) \quad \text{\textit{porta de esquecimento}};
\end{equation*}
\begin{equation}
\mathbf{c}_t = \mathbf{z}_t \odot \mathbf{i}_t + \mathbf{c}_{t-1} \odot \mathbf{f}_t \quad \text{\textit{célula}};
\end{equation}
\begin{equation}
\bar{\mathbf{o}}_t = \mathbf{W}_o \mathbf{x}_t + \mathbf{R}_o \mathbf{y}_{t-1} + \mathbf{b}_o,
\end{equation}
\begin{equation*}
\mathbf{o}_t = \sigma(\bar{\mathbf{o}}_t) \quad \text{\textit{porta de saída}};
\end{equation*}
\begin{equation}
\mathbf{y}_t = h(\mathbf{c}_t) \odot \mathbf{o}_t \quad \text{\textit{saída do bloco}}.
\end{equation}

Em que $\sigma$, $g$ e $h$ são funções de ativação não lineares ponto a ponto. A função sigmoide ($\sigma(x) = \frac{1}{1+e^{-x}}$) é usada como função de ativação da porta, e a tangente hiperbólica ($g(x) = h(x) = \tanh(x)$) é comumente usada como função de ativação de entrada e saída do bloco. A multiplicação ponto a ponto de dois vetores é denotada por $\odot$.

\subsection{Unidade GRU}
\label{sec:teoriaGRU}
As Unidades Recorrentes com Portas \textit{Gated Recurrent Units} (GRU), introduzidas por \cite{JunyoungChung2014}, são uma variação das LSTM. Enquanto as LSTM possuem três portões e uma célula de memória, as GRU simplificam essa estrutura ao fundir os portões de entrada e esquecimento em um único portão de atualização. Essa simplificação tem como objetivo tornar o treinamento mais eficiente e reduzir o número de parâmetros, mantendo um desempenho comparável às LSTM.

As fórmulas vetoriais para uma passagem direta em uma camada GRU foram encontradas no artigo de \cite{Cheng2024} de uma forma mais simplista que são:

\begin{equation}
	\bar{\mathbf{z}}_t = \mathbf{W}_{i\bar{z}} \mathbf{x}_t + \mathbf{b}_{i\bar{z}} + \mathbf{R}_{y\bar{z}} \mathbf{y}_{t-1} + \mathbf{b}_{y\bar{z}},
\end{equation}
\begin{equation*}
	\mathbf{z}_t = \sigma(\bar{\mathbf{z}}_t) \quad \text{\textit{portão de atualização}};
\end{equation*}
\begin{equation}
	\bar{\mathbf{r}}_t = \mathbf{W}_{i\bar{r}} \mathbf{x}_t + \mathbf{b}_{i\bar{r}} + \mathbf{R}_{y\bar{r}} \mathbf{y}_{t-1} + \mathbf{b}_{y\bar{r}},
\end{equation}
\begin{equation*}
	\mathbf{r}_t = \sigma(\bar{\mathbf{r}}_t) \quad \text{\textit{portão de reset}};
\end{equation*}
\begin{equation}
	\mathbf{\tilde{c}}_t = \mathbf{W}_{i\tilde{c}} \mathbf{x}_t + \mathbf{b}_{i\tilde{c}} + \mathbf{R}_{y\tilde{c}} (\mathbf{r}_t \odot \mathbf{y}_{t-1}) + \mathbf{b}_{y\tilde{c}},
\end{equation}
\begin{equation}
	\mathbf{c}_t = \tanh(\mathbf{\tilde{c}}_t) \quad \text{\textit{estado oculto}},
\end{equation}
\begin{equation}
	\mathbf{y}_t = \mathbf{z}_t \odot \mathbf{y}_{t-1} + (1 - \mathbf{z}_t) \odot \mathbf{c}_t .
\end{equation}

em que \textbf{W} e \textbf{R} são matrizes de pesos; \textbf{b} são vetores de viés; $\sigma$ é a função de ativação sigmoide e o $\odot$ denota a multiplicação ponto a ponto.

\begin{figure}[H]
\centering
\caption{Esquema detalhado da Unidade GRU}
\includegraphics[width=0.8\linewidth]{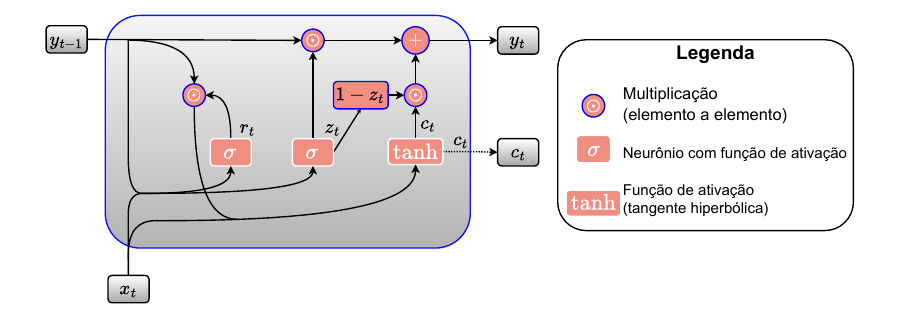}
\parbox{\linewidth}{\centering\noindent{Fonte: Elaborado pelo autor, adaptado de \cite{Cheng2024}.}}
\label{fig:GRU}
\end{figure}

A figura \ref{fig:GRU} mostra um esquema das Unidades Recorrentes com Portas (GRU) e a arquitetura típica dessa rede.

\subsection{Funções de Ativação}

\begin{figure}[H] 
\centering
\caption{Ilustração da transformação linear e da ativação linear em uma camada densa.}
\includegraphics[width=0.9\linewidth]{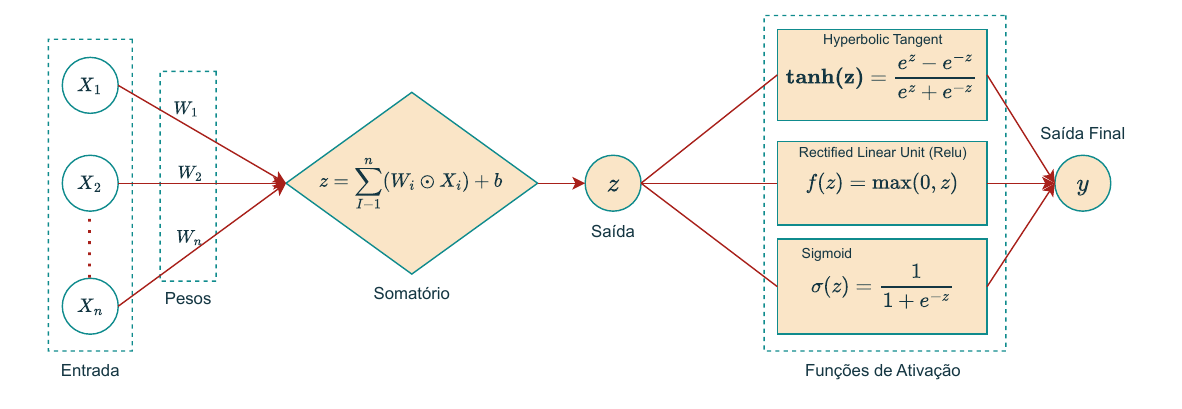}
\parbox{\linewidth}{\centering
\noindent{Fonte: Elaborado pelo autor}}
\label{fig:ativacao}
\end{figure}

As funções de ativação são componentes fundamentais em redes neurais, responsáveis por introduzir não-linearidades nas saídas das camadas, o que permite às redes neurais aprender e modelar relações complexas nos dados. Essas funções não possuem parâmetros ajustáveis e são fixas, usadas especificamente para introduzir não-linearidade nas redes neurais conforme \cite{Goodfellow2016}. A Figura \ref{fig:ativacao} ilustra a transformação linear e a ativação linear em uma camada densa final de uma rede neural.

\subsection{Entendendo o Funcionamento das Camadas Densas em Redes Neurais Recorrentes}
\label{sec:camadasDensas}
Os neurônios em redes neurais recorrentes (RNNs) são unidades fundamentais que processam informações ao longo do tempo. Eles são responsáveis por realizar operações matemáticas nos dados de entrada e nos estados ocultos anteriores (previsão do bloco anterior) para gerar saídas e atualizar seus próprios estados. Uma camada densa é uma camada comumente usada em redes neurais, em que cada neurônio na camada está totalmente conectado a todos os neurônios na camada anterior. Os cálculos realizados em uma camada densa envolvem multiplicação de matriz entre a entrada dos dados e os pesos (parâmetros) da camada, seguida por uma função de ativação. Aqui estão os cálculos para uma camada densa: Seja $\mathbf{x}$ a matriz de entrada de dimensão $(m \times n)$, em que $m$ é o número de amostras e $n$ é o número de características. Seja $\mathbf{W}$ a matriz de pesos da camada densa de dimensão $(n \times p)$, com $p$ sendo o número de neurônios na camada densa. Além disso, seja $\mathbf{b}$ o vetor de viés da camada densa de dimensão $(p \times n)$. 
A saída da camada densa $\mathbf{Z}$ é calculada da seguinte forma:
\[
\mathbf{Z} = \mathbf{xW} + \mathbf{b},
\]
aqui, $\mathbf{xW}$ representa a multiplicação de matriz entre a entrada e os pesos da camada densa, e $\mathbf{b}$ é o viés adicionado para produzir a saída final. 
É importante notar que após essa operação, geralmente é aplicada uma função de ativação aos elementos de $\mathbf{Z}$ para introduzir não linearidade na camada densa conforme \cite{Goodfellow2016}.

\subsection{Algoritmo de Otimização Adam: Uma Visão Geral}

O algoritmo Adam, desenvolvido por \cite{Kingma2014}, utiliza médias móveis exponenciais dos gradientes para atualizar os parâmetros, acelerando a convergência e evitando que o modelo fique preso em mínimos locais. O Adam incorpora estimativas de primeira e segunda ordens com correções de viés para melhorar a eficácia da otimização.
As configurações padrão para os problemas de aprendizado de máquina testados são $\alpha = 0.001$, $\beta_{1} = 0.9$, $\beta_{2} = 0.999$ e $\epsilon = 10^{-8}$. Todas as operações em vetores são realizadas elemento a elemento (matricialmente). Com $\beta_{t}^{1}$ e $\beta_{t}^{2}$ denotados como $\beta_{1}$ e $\beta_{2}$ elevados à potência $t$.
\begin{algorithm}[H]
\caption{: Algoritmo de Otimização Adam}

$\alpha$: Tamanho do Passo (Padrão sugerido: 0.001); 
$\beta_1, \beta_2 \in [0, 1)$: Taxas de Decaimento Exponencial para as Estimativas de Momento; 
$f(\theta)$: Função Objetiva Estocástica com Parâmetros $\theta$; 
$g_t$: É o vetor dos gradientes da função objetiva estocástica $f_t(\theta)$ em relação aos parâmetros $\theta$, calculado no passo $t$;
$\theta_0$: Vetor de Parâmetros Inicial. 
Inicialização:
$m_0 \leftarrow 0$ (Inicializar 1º Momento);
$v_0 \leftarrow 0$ (Inicializar 2º Momento);
$t \leftarrow 0$ (Inicializar Passo).\;
\While{$\theta_t$ não convergiu}{
  $t \leftarrow t + 1$;
  $g_t \leftarrow \nabla_{\theta}f_t(\theta_{t-1})$ (Obter gradientes em relação à função objetiva estocástica no passo $t$);\\
  $m_t \leftarrow \beta_1 \times m_{t-1} + (1 - \beta_1) \times g_t$ (Atualizar estimativa viesada do 1º Momento);\\
  $v_t \leftarrow \beta_2 \times v_{t-1} + (1 - \beta_2) \times g_t \times g_t$ (Atualizar estimativa viesada do 2º Momento);\\
  $\hat{m}_t \leftarrow \frac{m_t}{1 - \beta_1^t}$ (Calcular estimativa corrigida de viés do 1º Momento);\\
  $\hat{v}_t \leftarrow \frac{v_t}{1 - \beta_2^t}$ (Calcular estimativa corrigida de viés do 2º Momento);\\
  $\theta_t \leftarrow \theta_{t-1} - \alpha \times \frac{\hat{m}_t}{\sqrt{\hat{v}_t} + \epsilon}$ (Atualizar parâmetros);\\
  \textbf{end while}
}
\textbf{return} $\theta_t$ (Parâmetros Resultantes)\;

\end{algorithm}

\subsubsection{Descrição dos Parâmetros}

De acordo com \cite{Kingma2014}, o algoritmo Adam é considerado uma técnica avançadda de otimização que calcula taxas de aprendizado adaptativas para cada parâmetro. Ele combina características dos métodos Adagrad e RMSprop, mantendo médias móveis exponenciais dos gradientes e dos gradientes ao quadrado para ajustar as taxas de aprendizado.

\begin{flushright}
	\begin{minipage}{0.85\textwidth} %
\footnotesize
\rightskip=0pt plus 1fil 
\leftskip=0pt 
\justifying{$m_t$ e $v_t$ são estimativas do primeiro momento (a média) e do segundo momento (a variância não centralizada) dos gradientes, respectivamente, daí o nome do método. Como $m_t$ e $v_t$ são inicializados como vetores de zeros, os autores do Adam observam que eles são tendenciosos a valores próximos de zero, especialmente durante os passos iniciais, e particularmente quando as taxas de decaimento são pequenas [...].} \cite[p. 7]{Ruder2016}.
\end{minipage}
\end{flushright}

\footnote{Texto original: \enquote{$m_t$ and $v_t$ are estimates of the first moment (the mean) and the second moment (the uncentered variance) of the gradients respectively, hence the name of the method. As $m_t$ and $v_t$ are initialized as vectors of 0’s, the authors of Adam observe that they are biased towards zero, especially during the initial time steps, and especially when the decay rates are small.} (Ruder, 2016).}

Eles contrabalançam esses vieses calculando estimativas corrigidas de viés para o primeiro e segundo momentos:

\begin{equation}
m_t = \beta_1 m_{t-1} + (1 - \beta_1) g_t
\end{equation}

\begin{equation}
v_t = \beta_2 v_{t-1} + (1 - \beta_2) g_t^2
\end{equation}

Aqui, $m_t$ representa a estimativa da média dos gradientes e $v_t$ a estimativa da variância não centralizada. Para corrigir o viés de inicialização dessas estimativas, são calculadas as correções de viés:

\begin{equation}
\hat{m}_t = \frac{m_t}{1 - \beta_1^t}
\end{equation}

\begin{equation}
\hat{v}_t = \frac{v_t}{1 - \beta_2^t}
\end{equation}

Com essas estimativas corrigidas, a atualização dos parâmetros é dada por:

\begin{equation}
\theta_{t+1} = \theta_t - \frac{\eta}{\sqrt{\hat{v}_t} + \epsilon} \hat{m}_t
\end{equation}

Reiterando o que foi afirmado no início desta seção, os valores padrão sugeridos para os hiperparâmetros são $\beta_1 = 0.9$, $\beta_2 = 0.999$, e $\epsilon = 10^{-8}$. O otimizador Adam é conhecido por sua eficácia em uma ampla gama de problemas de aprendizado de máquina, proporcionando uma atualização eficiente e eficaz dos parâmetros durante o treinamento de redes neurais.

\section{Metodologia}
\label{sec:metodologia}

Nesta seção, descrevemos o procedimento adotado para modelar e prever séries temporais utilizando redes neurais recorrentes. Utilizamos dados de contagem dos focos ativos detectados pelo satélite \textit{AQUA\_M-T} no bioma da Amazônia, abrangendo uma série histórica registrada desde junho de 1998 até 31 de agosto de 2024. Esses dados estão disponíveis no \cite{INPE2024}. O processo metodológico para modelar e prever essa série temporal segue práticas estabelecidas na literatura de séries temporais e aprendizado de máquina. Inicialmente, dividimos os dados em conjuntos de treino e teste para avaliar a performance do modelo, aplicando técnicas como validação cruzada para assegurar a robustez do modelo. Após garantir que o modelo apresentava uma boa capacidade de generalização, optamos por treinar o modelo final utilizando 100\% dos dados disponíveis. Essa abordagem visa maximizar a precisão das previsões, especialmente em cenários de passos à frente da última observação treinada, como indicado por \cite{Geron2017}. No contexto de \textit{deep learning}, onde ajustes finos (\textit{fine-tuning}) são comuns, o uso do conjunto completo de dados após validação é uma prática justificada para aprimorar o desempenho, conforme discutido por \cite{Goodfellow2016}. Dessa forma, utilizamos o modelo treinado com 100\% dos dados para realizar previsões de \(n\) passos à frente, garantindo que as previsões fossem baseadas na maior quantidade de informações possível.

\subsection{Preparação dos Dados}

Na preparação dos dados, adotamos uma abordagem de treino, validação e teste adaptada para séries temporais contínuas. Para garantir a eficácia da avaliação do modelo, seguimos o processo de divisão dos dados de frente para trás. Primeiramente, removemos os últimos 12 \textit{lags} (meses) da série para o conjunto de teste, considerando a série completa menos esses 12 \textit{lags}. Em seguida, removemos 24 \textit{lags} adicionais para o conjunto de validação, o que deixou a série completa menos 36 \textit{lags} para o treinamento. Embora tenhamos seguido a abordagem de divisão de dados, utilizamos validação cruzada com duas sementes para avaliar a performance do modelo. Conforme descrito por \cite{Geron2017}, a validação cruzada é essencial para garantir que o modelo generalize bem para novos dados. Foram utilizadas duas sementes distintas para criar dois modelos diferentes, o que permitiu avaliar a robustez e a estabilidade do modelo. A divisão final dos dados foi a seguinte:
\begin{itemize}
  \item \textbf{Conjunto de Treino}: Junho de 1998 até agosto de 2021;
  \item \textbf{Conjunto de Validação}: Setembro de 2021 até agosto de 2023;
  \item \textbf{Conjunto de Teste}: Setembro de 2023 até agosto de 2024.
\end{itemize}
Essa abordagem, combinada com o uso de sementes fixas, garantiu a replicabilidade dos resultados e confirmou a consistência das generalizações para os conjuntos de treino, validação e teste.

\subsection{Configuração dos Modelos}
Foi utilizado a combinação dos modelos de redes neurais recorrentes LSTM+GRU essa arquitetura consiste em:

\begin{itemize}
  \item \textbf{Camada de Entrada}: Recebe os dados da série temporal em janelas fixadas previamente, que na nossa arquitetura escolhemos tamanho de 12 para que a partir de 12 meses se tenha a primeira previsão no 13° ponto.
  \item \textbf{Camada Recorrente}: Para o modelo LSTM+GRU, foi configurada uma camada LSTM seguida por uma camada GRU, ambas com 256 neurônios.
  \item \textbf{Camada Densa}: Uma camada densa com 256 neurônios e função de ativação ReLU.
  \item \textbf{Camada de Saída}: Uma camada densa com 1 neurônio e ativação linear, fornecendo a previsão final para cada janela de entrada.
\end{itemize}

\begin{figure}[H]  
\centering
\caption{Esquema detalhado da Unidade GRU e camadas densas na saída final (previsão) em uma rede LSTM.}
\includegraphics[width=0.9\linewidth]{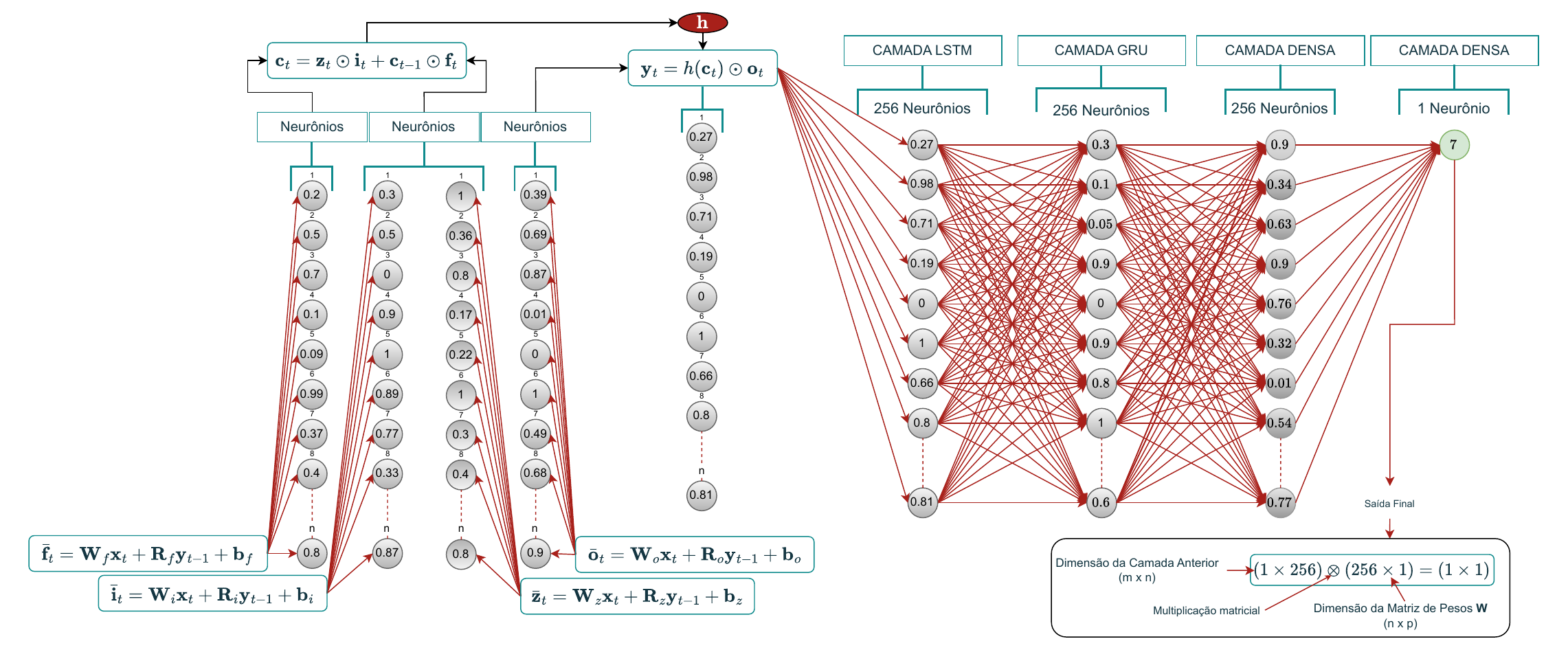}
\parbox{\linewidth}{\centering
\noindent{Fonte: Elaborado pelo autor}}
\label{fig:camadasDetalhe}
\end{figure}

 Veja a figura \ref{fig:camadasDetalhe} que melhor ilustra essa configuração.

A Figura \ref{fig:camadasDetalhe} ilustra uma arquitetura de rede neural que inclui as seguintes camadas:
\begin{itemize}
\item Camada LSTM com 256 neurônios;
\item Camada GRU com 256 neurônios;
\item Camada densa com 256 neurônios;
\item Camada densa de saída com 1 neurônio.
\end{itemize}
A figura \ref{fig:camadasDetalhe} ilustra a transmissão de informações entre as camadas até a saída final e não aborda o funcionamento de dropout ou funções de ativação. A explicação da arquitetura é a seguinte: 

Cada entrada \( X_t \) no tempo \( t \) é inicialmente processada pela camada LSTM composta por 256 neurônios. A saída dessa camada LSTM, com dimensão \( (256 \times 1) \) — considerando que nosso trabalho envolve uma única variável ao longo do tempo (focos ativos) — resulta em um vetor de dimensão \( (256 \times 1) \). Essa saída é então utilizada como entrada para a camada GRU, que também possui 256 neurônios. A saída da camada GRU é processada por outra camada densa com 256 neurônios, mantendo a dimensão \( (256 \times 1) \). Finalmente, essa saída é alimentada na camada densa de saída com 1 neurônio, resultando em uma previsão única para o próximo ponto da série temporal. Para ilustrar o funcionamento, considere um bloco de dados com 12 valores, em que a entrada \( X_t \) é o vetor de valores de 1 a 12. A previsão é feita para o 13º valor. Esse processo é repetido ao mover a janela de entrada, de modo que o segundo bloco será de 2 a 13, e a previsão será para o 14º ponto, e assim por diante até o final da série. Esta ilustração é de apenas um bloco, mas ao configurar a série com blocos de tamanho 12, o modelo recebe os dados configurados previamente. Assim, se eu tenho valores de 1 a \( n \), os dados são configurados em \textit{arrays} de tamanho 12, sendo \( X_{t1} \) de 1 a 12, \( X_{t2} \) de 2 a 13, e assim por diante até o final da série. Portanto, essa ilustração é de apenas um bloco, mas no funcionamento completo da rede, o modelo computa todos os blocos pré-definidos e resulta em previsões para cada \( X_{ti} \) de acordo com o tamanho da série.

\subsection{Treinamento e Avaliação dos Modelos}
\label{sec:treinamentoAvaliacaoModelos}
Os modelos foram treinados utilizando a linguagem de programação \cite{Python2024}, com as bibliotecas \textit{Scikit-learn} \cite{ScikitLearn2011} e \cite{FrançoisChollet2024}. Cada modelo foi treinado por 1000 épocas, e as métricas Erro Absoluto Médio (MAE) e Raiz do Erro Quadrático Médio (RMSE) foram utilizadas para avaliar o desempenho.
Para cada época, o modelo gerou previsões utilizando uma técnica conhecida como janela/bloco deslizante (ou \textit{sliding window}). Este procedimento funciona da seguinte forma:

1. \textbf{Janela de Entrada}: Definimos uma janela de 12 \textit{lags}, ou seja, o modelo usa os dados dos 12 períodos anteriores para prever o valor do próximo período. Por exemplo, se temos dados mensais e a janela é de 12 \textit{lags}, o modelo usa os dados dos últimos 12 meses para prever o valor do mês seguinte.
2. \textbf{Deslizamento da Janela}: Após gerar uma previsão para o próximo período (o 13º), a janela é deslocada uma posição para frente. Isso significa que a previsão é feita usando os dados dos períodos de 2 a 13 para prever o 14º período. Esse processo é repetido até que todas as previsões sejam feitas para o restante da série temporal.
3. \textbf{Avaliação das Previsões}: As previsões geradas para cada época são comparadas com os dados reais da série temporal. Para avaliar a precisão das previsões, utilizamos as métricas MAE e RMSE. O MAE calcula a média dos erros absolutos das previsões, enquanto o RMSE calcula a raiz quadrada da média dos erros quadráticos. Os parâmetros iniciais dos modelos foram definidos utilizando uma distribuição de probabilidade específica, a distribuição normal de He (He normal). Essa distribuição é definida com média zero e desvio padrão \(\sqrt{2 / n}\), onde \(n\) é o número de unidades na camada de entrada. A escolha dessa distribuição ajuda a garantir uma inicialização adequada dos pesos, facilitando o treinamento eficaz de redes neurais profundas, conforme os desenvolvedores \cite{KerasDevelopers2024}.

Os dados foram divididos da seguinte forma:
\begin{itemize}
  \item \textbf{Treino}: Junho de 1998 até agosto de 2021;
  \item \textbf{Validação}: Setembro de 2021 até agosto de 2023;
  \item \textbf{Teste}: Setembro de 2023 até agosto de 2024.
\end{itemize}
Utilizamos duas sementes distintas para garantir a replicabilidade dos resultados e a estabilidade das métricas de desempenho. O modelo que apresentou o menor erro médio nas métricas MAE e RMSE durante o treinamento foi selecionado como o modelo final.

\subsection{Como as Previsões são Calculadas?}
\label{sec:previsoes}

As previsões foram realizadas após o treinamento completo dos dados de focos ativos registrados na região da Amazônia, Brasil, disponíveis no \cite{INPE2024}, abrangendo o período de junho de 1998 até agosto de 2024. Após o treinamento, utilizamos a função \textit{predict} do pacote \cite{FrançoisChollet2024}, para gerar as previsões. O processo envolveu o uso do modelo treinado, que já contém todos os parâmetros otimizados e ajustados. O modelo, armazenado e salvo como o melhor obtido durante o treinamento, é utilizado com a função \textit{predict}, que é chamada como \textit{modelo.predict()}. Este modelo foi treinado com uma variável e um bloco de tamanho 12. A função \textit{predict} segue a sequência dos dados, incorporando as previsões anteriores para gerar novos resultados, adicionando esses resultados à série temporal e prevendo o próximo ponto. Para detalhar o processo: a função \textit{predict} utiliza as últimas 12 observações da série temporal (o tamanho da janela deslizante), que vão de setembro de 2023 a agosto de 2024, para prever o 13º ponto, que corresponde a setembro de 2024. A abordagem de \enquote{janela deslizante} é usada, permitindo que após a primeira previsão para setembro de 2024, o modelo integre essa previsão e gere uma nova previsão para o próximo mês. No segundo passo, por exemplo, ele utiliza as observações de outubro de 2023 a setembro de 2024, agora incluindo a previsão anterior (de setembro), para prever outubro de 2024 (um dado que ainda não existe na série temporal). No terceiro passo, o modelo usa os dados de novembro de 2023 a outubro de 2024, incluindo as previsões obtidas de setembro e outubro à série temporal, e assim por diante. Esse processo continua até que todas as previsões dos 12 meses sejam realizadas. Essa abordagem assegura que cada previsão mensal se baseie nos dados históricos mais recentes, juntamente com as previsões feitas nos passos anteriores, resultando em uma modelagem robusta para séries temporais de dados de contagem, conforme descrito na literatura e referenciado nesta seção. A ilustração detalhada desse processo está apresentada na Figura \ref{fig:camadasDetalhe}.

\section{Resultados da Análise Estatística}
Nesta seção, apresentamos uma análise descritiva da série temporal de focos ativos na Amazônia, com ênfase na média, desvio padrão, variância e nos valores máximos e mínimos registrados ao longo dos anos, veja a Tabela \ref{tab:descritiva}.

\begin{table}[H]
\centering
\caption{Estatísticas descritivas dos focos ativos na Amazônia}
\begin{tabular}{ccccc}
\hline
Mínimo & Média & Máximo & Desvio Padrão & Variância \\
\hline
70 & 9084,378 & 73141 & 12596,04 & 158660137 \\
\hline
\end{tabular}
\label{tab:descritiva}\\
\noindent Fonte: Autor, baseado nos dados da base de dados fornecida pelo \textit{Instituto Nacional de Pesquisas Espaciais (INPE)}
\end{table}

A análise foi realizada cuidadosamente, aplicando técnicas de análise de dados para identificar os pontos mais extremos de cada ano. Nosso objetivo é oferecer uma visão clara e direta desses valores, evitando a complexidade que seria introduzida por uma tabela detalhada. Ao invés disso, optamos por uma representação gráfica que facilita a visualização e compreensão dessas estatísticas importantes.

\begin{figure}[!ht]
\centering
\caption{Série temporal dos focos ativos na Amazônia (1998-2024)}
\includegraphics[width=0.9\linewidth]{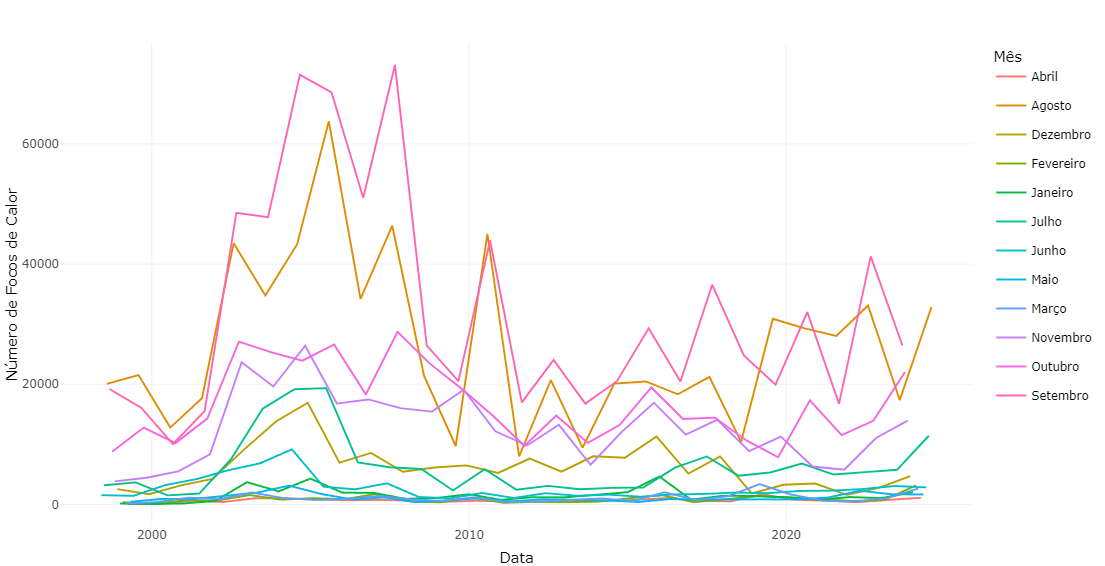}\\
\noindent{Fonte: Autor, baseado nos dados do \textit{Instituto Nacional de Pesquisas Espaciais (INPE)}. \label{fig:serieTemp}}
\end{figure}

A série temporal de junho de 1998 até agosto de 2024 apresenta dados mensais do total de focos ativos registrados pelo satélite de referência a cada mês. Como vemos na Figura \ref{fig:serieTemp}, os meses de agosto e setembro foram consistentemente registrados como aqueles com o maior número de focos ativos durante esse período de mais de 20 anos.

\begin{figure}[!ht]
\centering
\caption{Valores extremos dos focos ativos registrados na Amazônia no período de 1998 a 2024}
\includegraphics[width=0.9\linewidth]{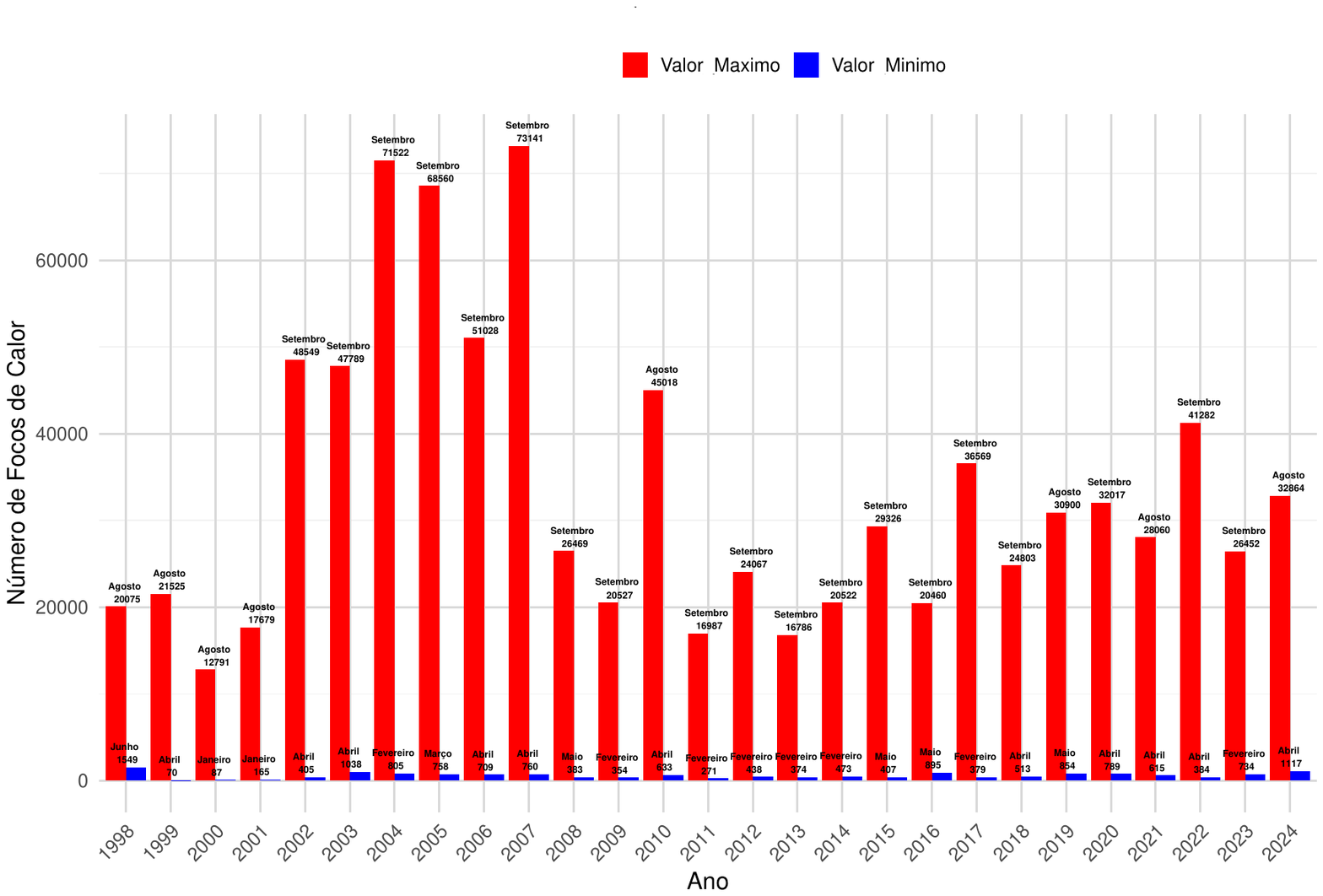}\\
\noindent{Fonte: Autor, baseado nos dados \textit{Instituto Nacional de Estudos e Pesquisas Espaciais (INPE)}. \label{pontosExtremos}}
\end{figure}

A figura \ref{pontosExtremos} apresenta os pontos extremos de focos ativos de cada ano, enfatizando a sazonalidade existente na série temporal da Amazônia. Desde 1998 até 2024, observa-se que os maiores índices de focos ativos ocorrem consistentemente nos meses de agosto e setembro, enquanto os menores índices são registrados no primeiro semestre, principalmente nos meses de janeiro, fevereiro, maio e abril.

\begin{figure}[H]
	\centering
	\caption{Escaneie o QR code ou acesse o link \href{https://newstatistic.shinyapps.io/InteractiveAmazonHeatmap/}{\texttt{https://newstatistic.shinyapps.io/InteractiveAmazonHeatmap/}}.}
	\includegraphics[width=0.2\linewidth]{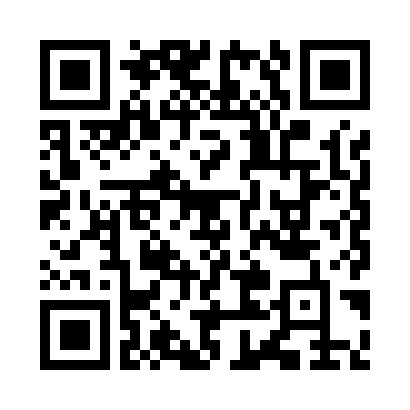}
	\\
	\noindent Fonte: Elaborado pelo autor
	\label{fig:qrCode}
\end{figure}

Para explorar gráficos detalhados e obter uma visualização interativa da série histórica de focos ativos na Amazônia, você pode escanear o QR code da Figura \ref{fig:qrCode}. Este QR Code direcionará para uma aplicação desenvolvida na linguagem de programação \cite{RCoreTeam2024} e \cite{shiny2024}.

\section{Resultados da Análise de Treinamento do Modelo de Aprendizado de Máquina}
Nesta seção, apresentamos e discutimos os resultados obtidos para os modelos de redes neurais recorrentes avaliados, especificamente a abordagem mista que combina LSTM e GRU. Vamos explorar o desempenho do modelo, utilizando métricas de avaliação, como Raiz do Erro Quadrático Médio (RMSE) e Erro Absoluto Médio (MAE), tanto para os conjuntos de treino quanto para os de teste. Cada modelo foi avaliado isoladamente, e os resultados obtidos serão apresentados em tabelas detalhadas. Utilizaremos essas métricas para comparar o desempenho dos modelos e determinar qual deles apresenta os menores valores de erro. O modelo que demonstrar melhor desempenho, com os menores valores de erro, será selecionado como o mais eficaz para a tarefa de previsão em questão. As implicações dos resultados serão discutidas, incluindo a análise da variação das métricas com diferentes sementes e configurações. Esta análise fornecerá uma visão abrangente da eficácia de cada modelo, permitindo a escolha do modelo mais adequado para realizar as previsões necessárias com base nos critérios estabelecidos.

\section{Resultados do Modelo LSTM+GRU}

\begin{figure}[H]
    \centering
    \caption{Comparação das métricas de desempenho para o modelo treinado com as sementes 2024 e 2025}
    \begin{minipage}{0.48\linewidth}
        \centering
        \includegraphics[width=\linewidth]{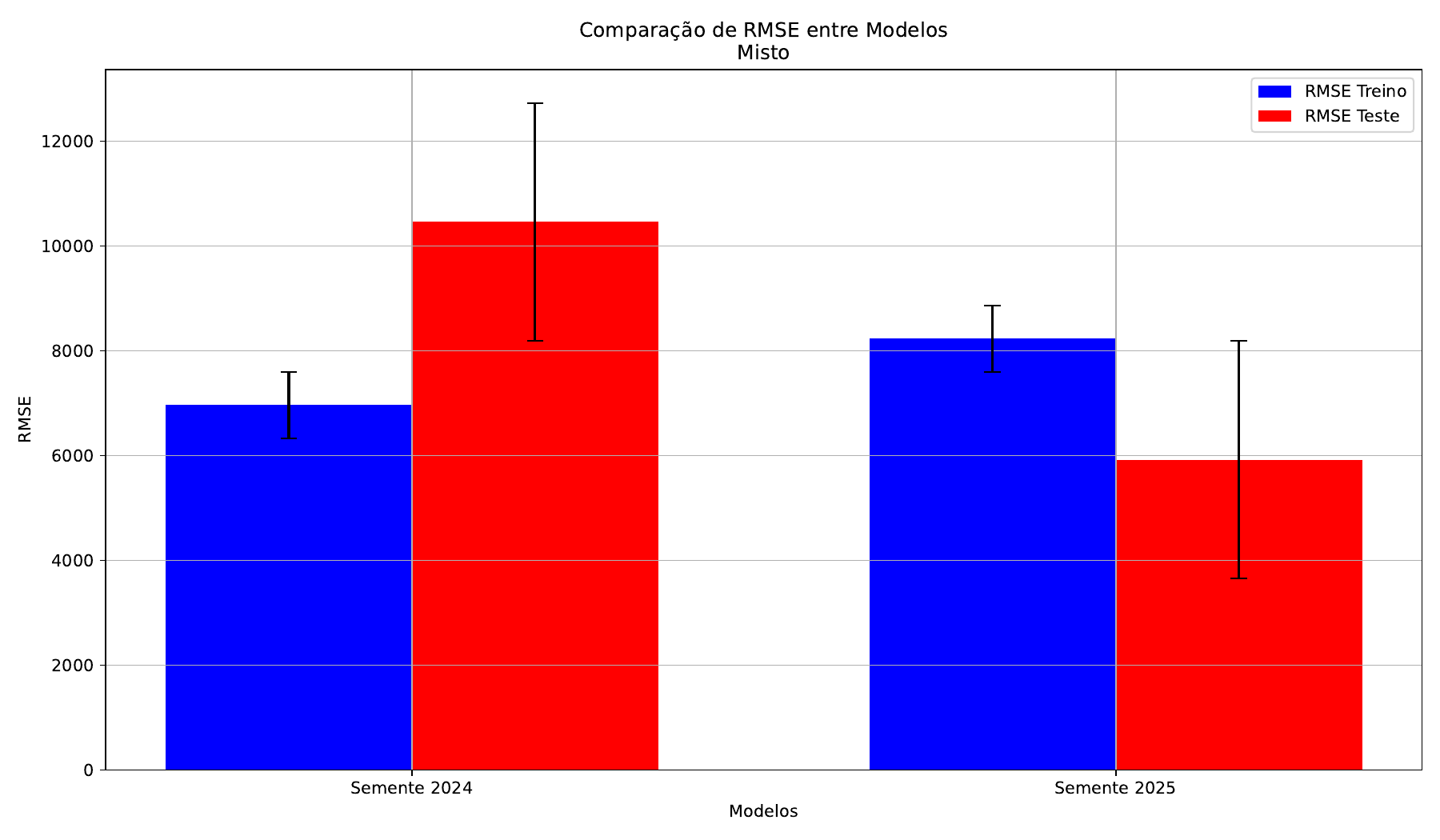}
        \caption*{Raiz do Erro Quadrático Médio (RMSE) - Tabela correspondente ao valor exato: (\ref{tab:LSTM_metrics}).}
    \end{minipage}
    \hfill
    \begin{minipage}{0.48\linewidth}
        \centering
        \includegraphics[width=\linewidth]{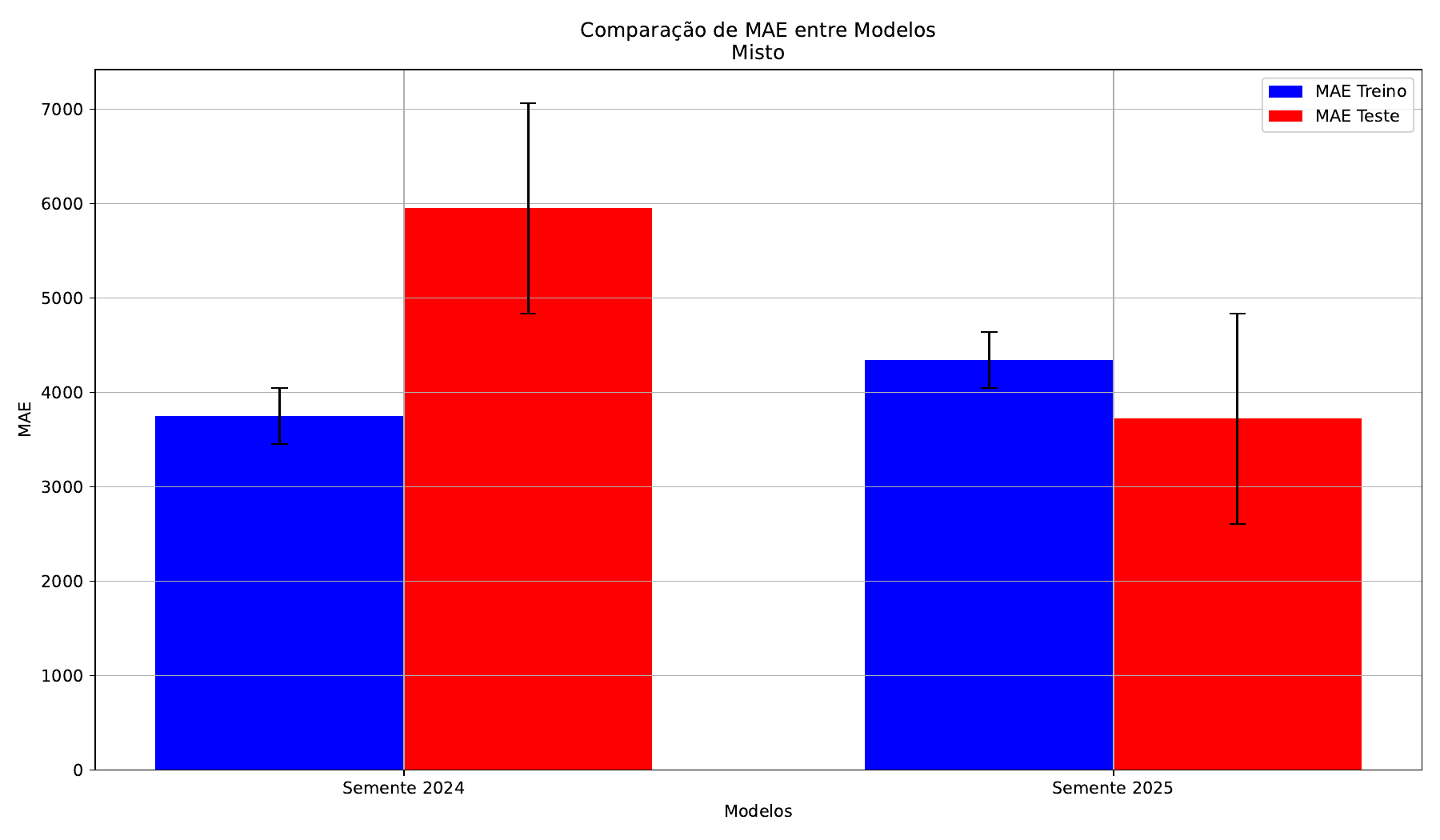}
        \caption*{Erro Albosoluto Médio (MAE) - Tabela correspondente ao valor exato: (\ref{tab:LSTM_metrics}).}
    \end{minipage}
    \caption*{Fonte: Autoria própria} 
    \label{fig:metricasTreinoTeste}
\end{figure}

A Tabela \ref{tab:LSTM_metrics} e a Figura \ref{fig:metricasTreinoTeste} relacionadas ilustram o desempenho do modelo LSTM+GRU para os conjuntos de treino e teste, utilizando diferentes sementes de inicialização. As métricas de Erro Quadrático Médio (RMSE) e Erro Absoluto Médio (MAE) são fundamentais para avaliar a precisão das previsões do modelo.

\begin{table}[H]
\centering
\caption{Métricas do modelo LSTM+GRU para os conjuntos treino e teste}
\begin{tabular}{ccccc}
\hline
Semente & RMSE Treino & RMSE Teste & MAE Treino & MAE Teste \\
\hline
2024 & 6900 & 10140 & 3790 & 5991 \\
2025 & 8079 & 5992 & 4138 & 3785 \\
\hline
\end{tabular}
\label{tab:LSTM_metrics}\\
\noindent  Fonte: Autor, baseado nos resultados das métricas
\end{table}

O Erro Absoluto Médio (MAE) mede a média das diferenças absolutas entre os valores reais e as previsões, e é definido pela seguinte fórmula:

\begin{equation}
\label{eq:mae}
\text{MAE} = \frac{1}{n} \sum_{i=1}^{n} \left| y_i - \hat{y}_i \right|,
\end{equation}

em que \( n \) é o número total de meses, \( y_i \) são os valores reais de cada mês, e \( \hat{y}_i \) são as previsões do modelo para cada mês. Nesse contexto, conseguimos obter para cada época todas as diferenças entre os valores mensais reais e o que o modelo LSTM+GRU prevê para cada um desses meses. Depois, extraímos a média dessas diferenças, que nada mais é do que a soma dessas diferenças absolutas dividida pelo total de meses (\( n \)). Já o Erro Quadrático Médio (RMSE) leva em consideração o quadrado dessas diferenças, penalizando erros maiores de forma mais severa, e é dado por:

\begin{equation}
\label{eq:rmse}
\text{RMSE} = \sqrt{\frac{1}{n} \sum_{i=1}^{n} \left( y_i - \hat{y}_i \right)^2},
\end{equation}

em que \( n \) é o número total de meses, \( y_i \) são os valores reais de cada mês, e \( \hat{y}_i \) são as previsões do modelo para cada mês. O RMSE calcula a raiz quadrada da média dos quadrados das diferenças entre os valores reais e as previsões. Isso penaliza erros maiores de forma mais intensa, fornecendo uma medida que reflete a magnitude dos erros em um nível mais severo do que o MAE.
Essas métricas são utilizadas para selecionar o melhor modelo entre os 1000 treinamentos realizados. Em cada época, a diferença entre os valores reais e as previsões é calculada, e o modelo que apresenta a menor diferença média é escolhido como o melhor. Observa-se que, embora haja uma diferença significativa nas previsões, especialmente no RMSE, o MAE nos fornece uma diferença média de pouco mais de 3700 focos ativos, em comparação a uma média histórica de 9000 focos. Isso sugere que, apesar de não ser extremamente preciso, o modelo ainda consegue capturar a tendência sazonal geral, com um erro que representa menos de 50\% da média histórica.

\begin{figure}[htbp]
	\centering
	\caption{Comparação dos conjuntos de treino e teste da validação cruzada fixadas nas sementes 2024 e 2025}
	\begin{subfigure}{0.48\textwidth}
		\centering
		\includegraphics[width=\linewidth]{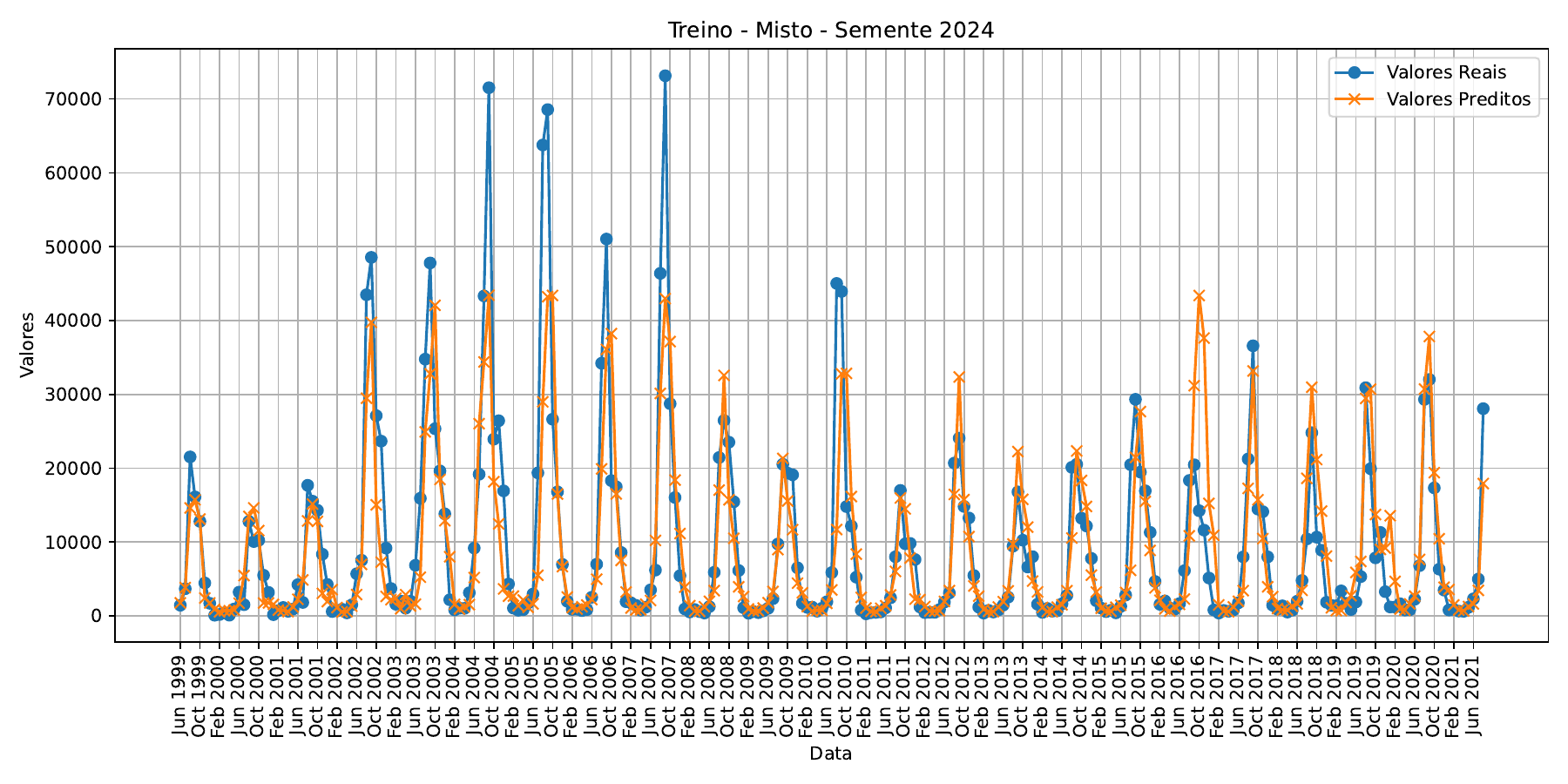}
		\caption*{Treinamento (semente 2024)}
		\label{fig:sub1}
	\end{subfigure}
	\hfill
	\begin{subfigure}{0.48\textwidth}
		\centering
		\includegraphics[width=\linewidth]{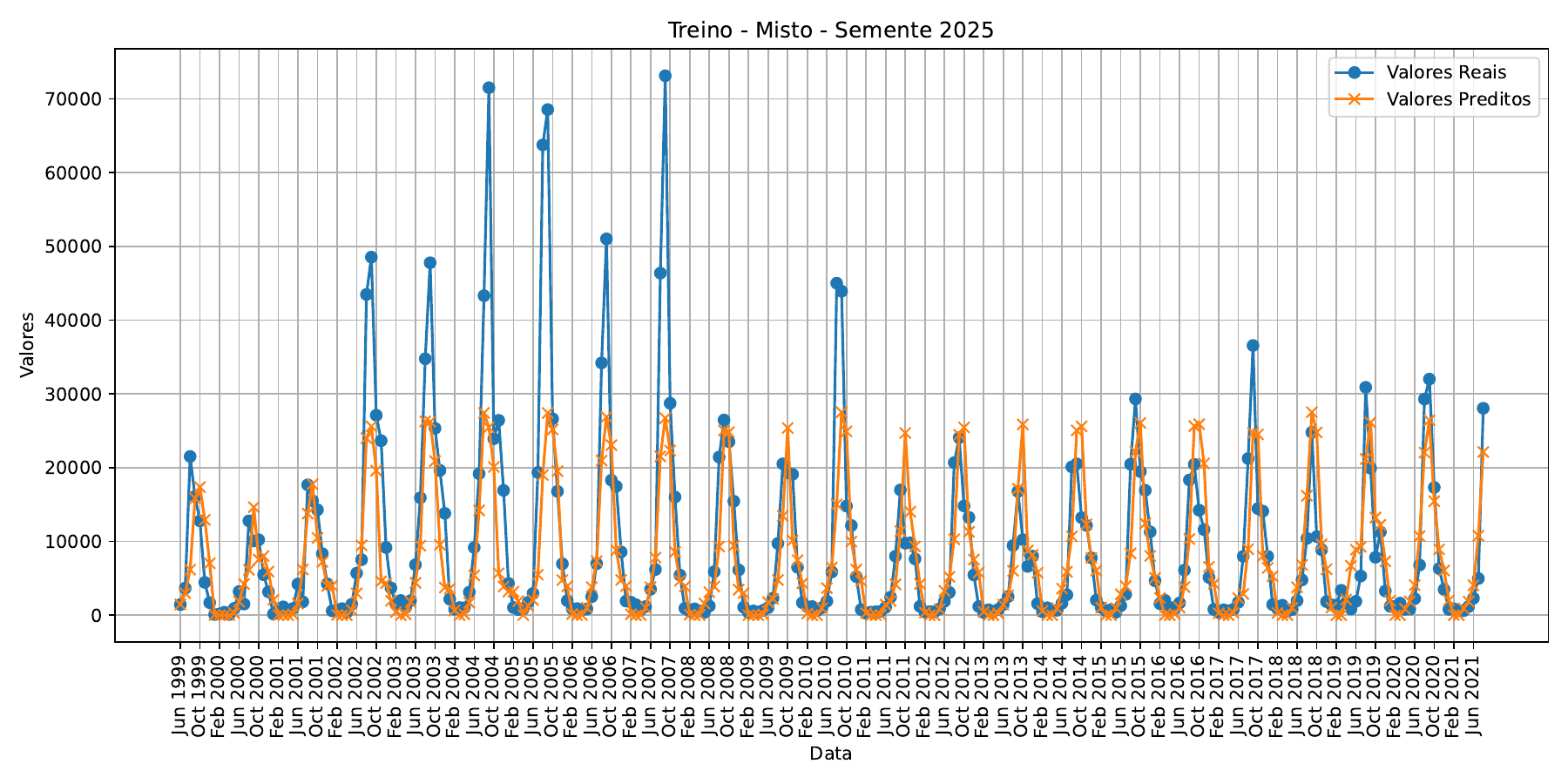}
		\caption*{Treinamento (semente 2025)}
		\label{fig:sub2}
	\end{subfigure}
	
	\begin{subfigure}{0.48\textwidth}
		\centering
		\includegraphics[width=\linewidth]{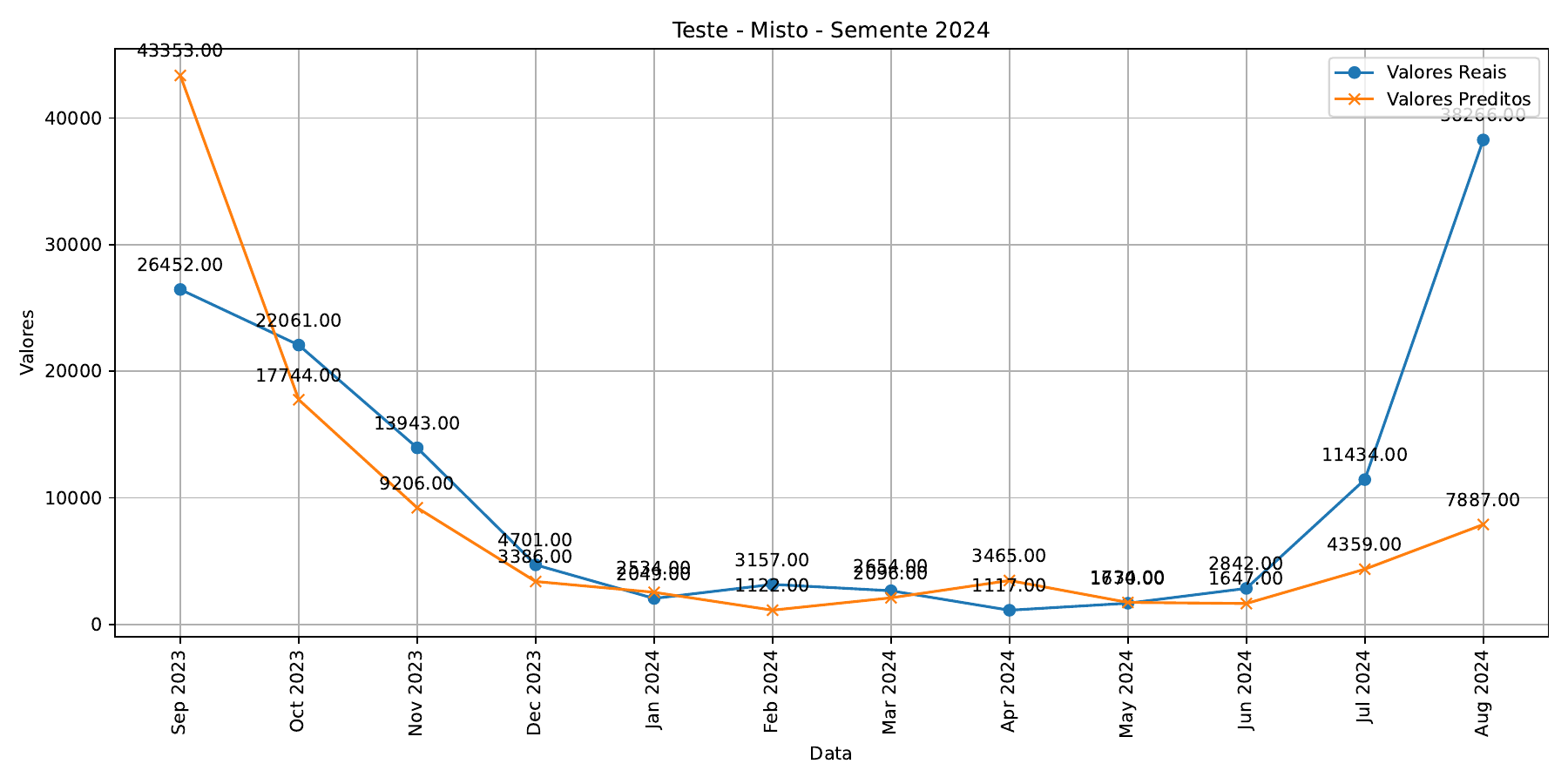}
		\caption*{Conjunto de Teste (Semente 2024)}
		\label{fig:sub3}
	\end{subfigure}
	\hfill
	\begin{subfigure}{0.48\textwidth}
		\centering
		\includegraphics[width=\linewidth]{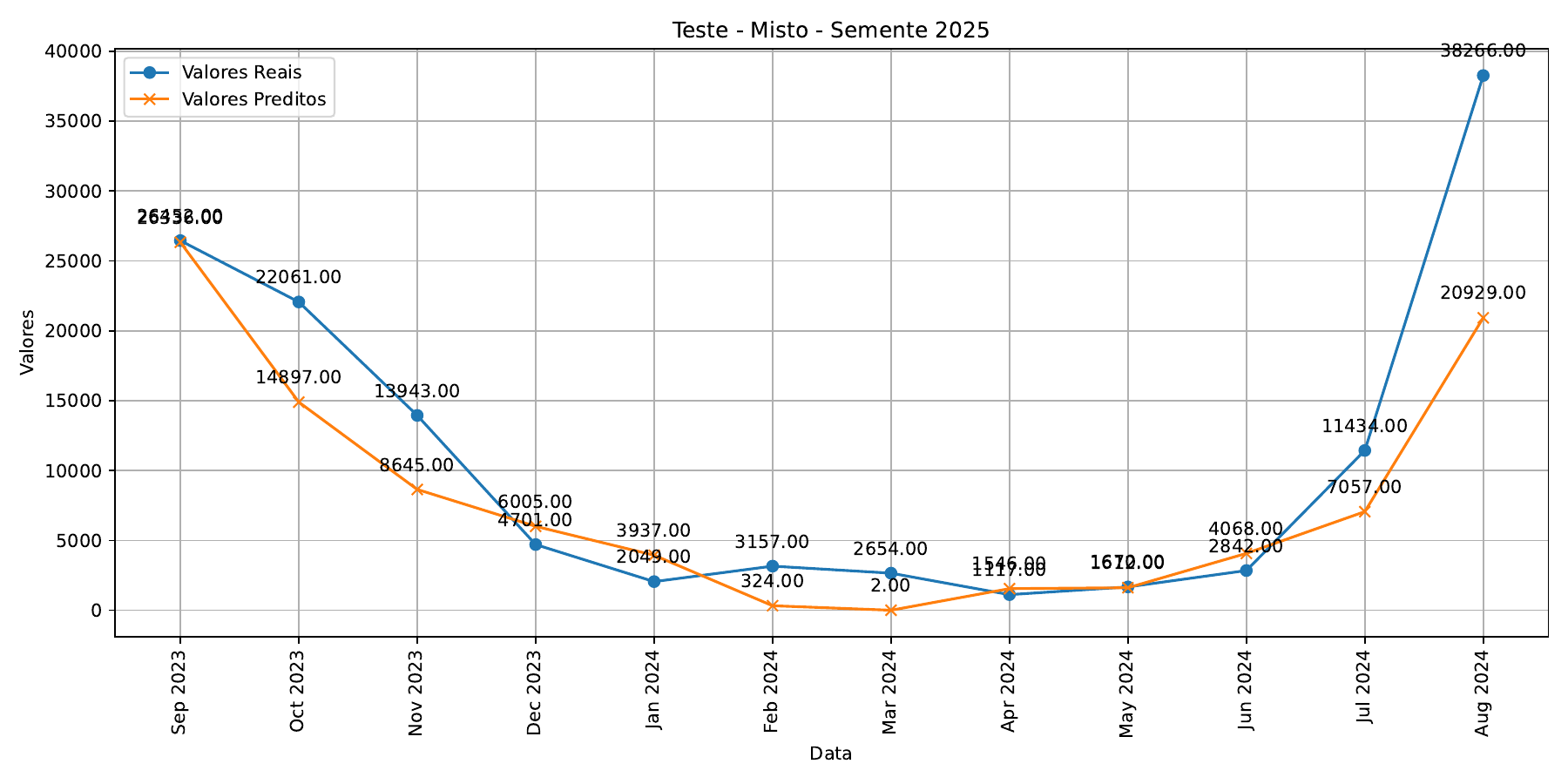}
		\caption*{Conjunto de Teste (Semente 2025)}
		\label{fig:sub4}
	\end{subfigure}
	\caption*{Fonte: Autoria própria} 
	\label{fig:imagensTreinoTeste}
\end{figure}

A Figura \ref{fig:imagensTreinoTeste} ilustra a validação cruzada realizada com duas sementes distintas, 2024 e 2025, comparando dois conjuntos de treino e teste. Esta abordagem é fundamental para avaliar a capacidade de generalização do modelo em séries temporais, onde a sequência dos dados é extremamente importante. Ao utilizar sementes diferentes para os conjuntos de treino, teste e validação, garantimos que a validação cruzada considere variações na inicialização do modelo e na estimação dos parâmetros, permitindo uma avaliação mais robusta da generalização.

\begin{figure}[H]
    \centering
    \caption{Comparação da perda (\textit{loss}) dos conjuntos de treino e validação, demonstrando o ponto exato dentre as 1000 épocas de treinamento em que cada um dos conjuntos obtiveram o menor Erro Absoluto Médio (MAE).}
    \begin{minipage}{0.48\linewidth}
        \centering
        \includegraphics[width=\linewidth]{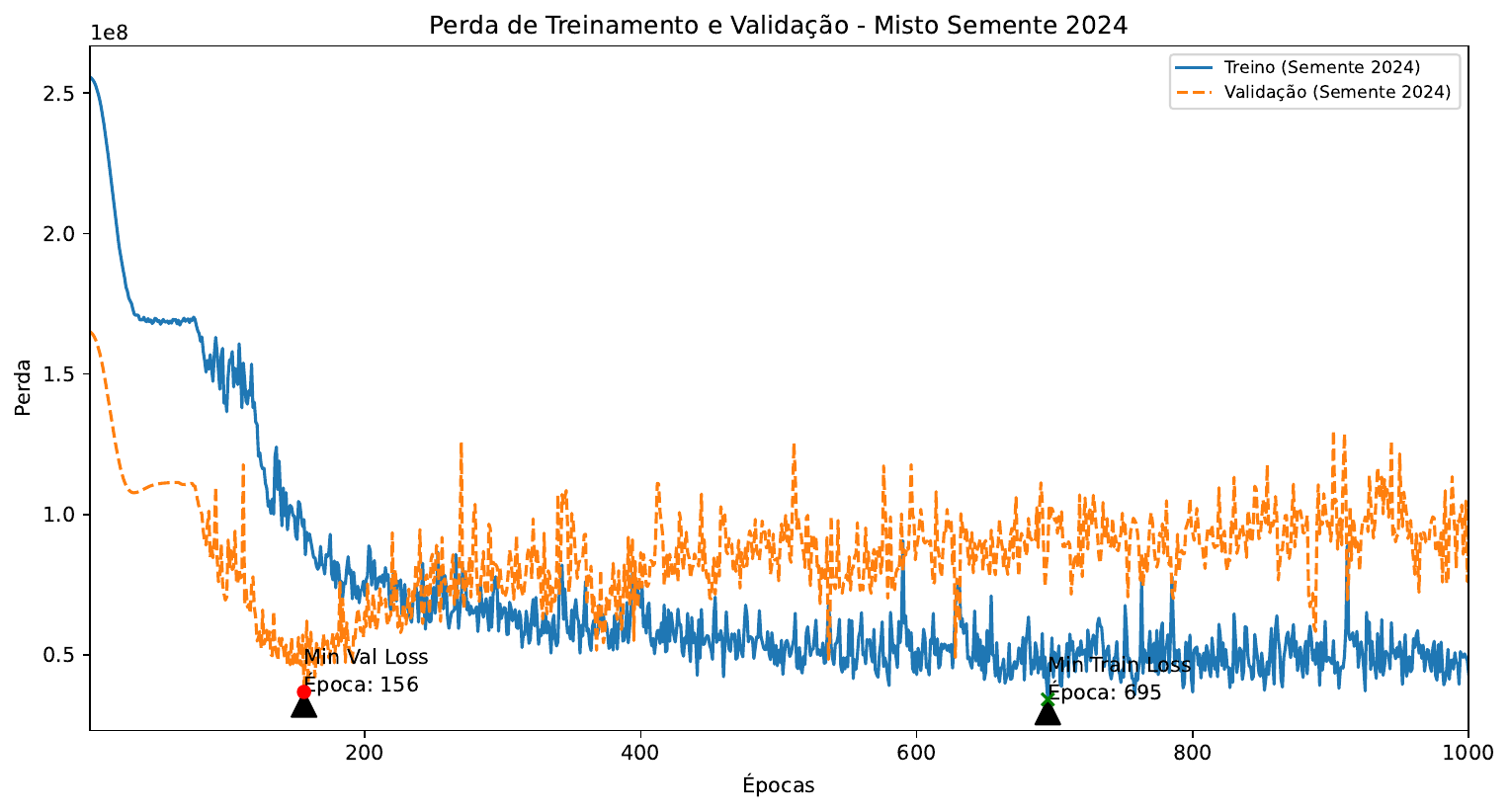}
        \caption*{Perda (\textit{Loss}) dos conjuntos de treino e validação (Semente 2024).}
    \end{minipage}
    \hfill
    \begin{minipage}{0.48\linewidth}
        \centering
        \includegraphics[width=\linewidth]{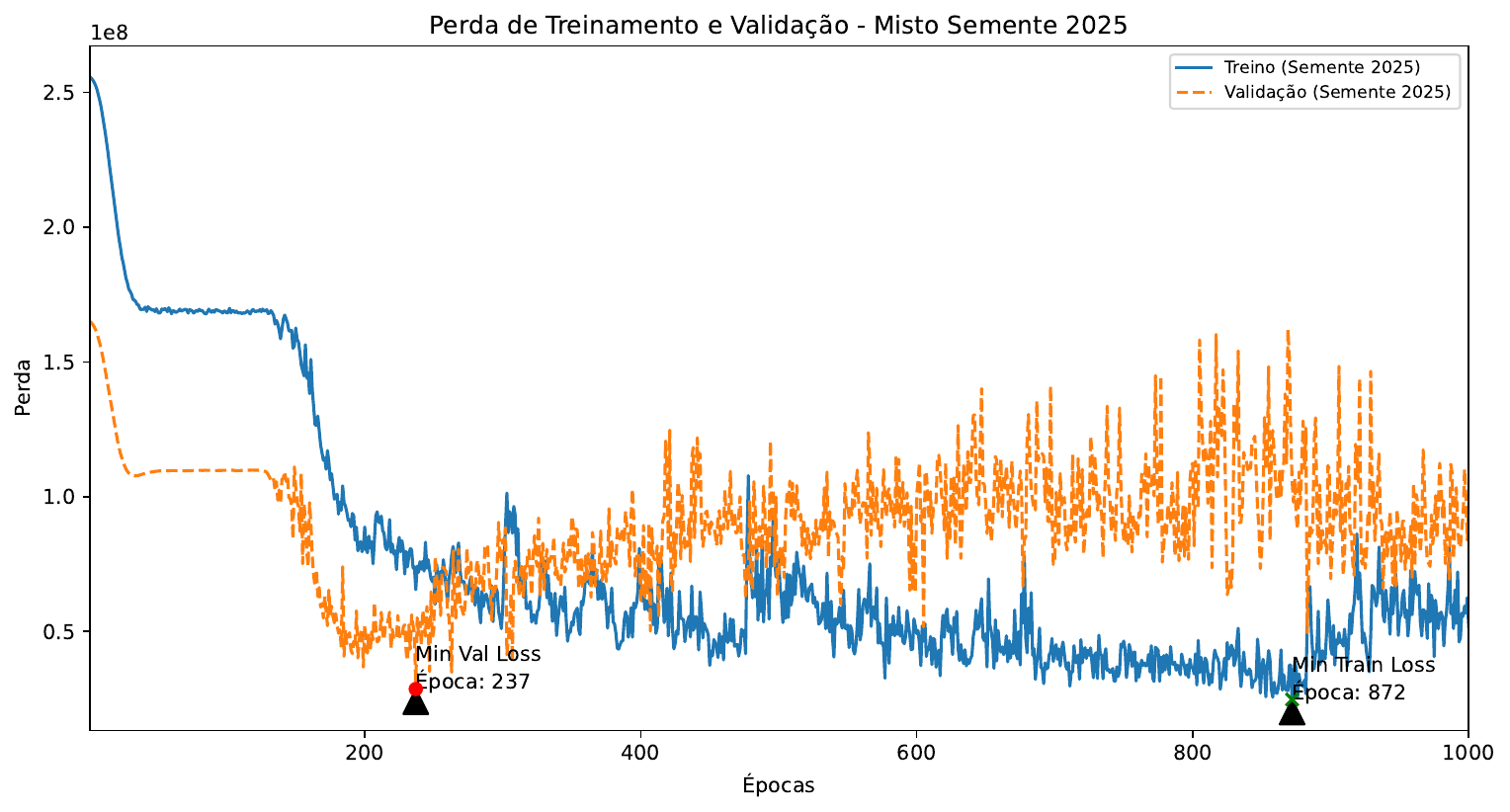}
        \caption*{Perda (\textit{Loss}) dos conjuntos de treino e validação (Semente 2025).}
    \end{minipage}
    \caption*{Fonte: Autoria própria} 
    \label{fig:imagensLoesVal}
\end{figure}

A Figura \ref{fig:imagensLoesVal} mostra a comparação da perda (\textit{Loss}) dos conjuntos de treinamento e validação para duas sementes diferentes: 2024 e 2025. A perda (\textit{Loss}) é uma métrica que representa o erro médio entre os valores reais e as previsões do modelo em cada época durante o treinamento. A fórmula da perda (\textit{Loss}) é diretamente relacionada às métricas de erro absoluto médio (MAE) e raiz do erro quadrático médio (RMSE), discutidas nas Equações \ref{eq:mae} e \ref{eq:rmse}. Esses gráficos são fundamentais para a análise do desempenho do modelo. A perda (\textit{Loss}) demonstra como os parâmetros do modelo são ajustados ao longo do tempo para minimizar o erro. O ponto onde a perda é minimizada indica a melhor configuração dos parâmetros do modelo para a previsão. A análise detalhada da perda nos conjuntos de treinamento e validação, conforme descrito na Seção \ref{sec:treinamentoAvaliacaoModelos}, revela a eficácia do ajuste do modelo. Observando a perda ao longo das 1000 épocas, é possível avaliar se o modelo está generalizando bem para novos dados, o que é essencial para prever a tendência dos dados. Portanto, esses gráficos ilustram a evolução da perda e fornecem detalhes sobre a capacidade da convergência dos parâmetros a cada modelo de se ajustar aos dados, refletindo diretamente na qualidade das previsões e na efetividade do treinamento realizado.

\subsection{Treinamento para os dados completos}

\begin{figure}[htbp]
\centering
\caption{Resultado do modelo treinado com todos os dados da série temporal que vai de junho de 1998 a agosto de 2024}

\begin{subfigure}{0.48\textwidth}
  \centering
  \includegraphics[width=\linewidth]{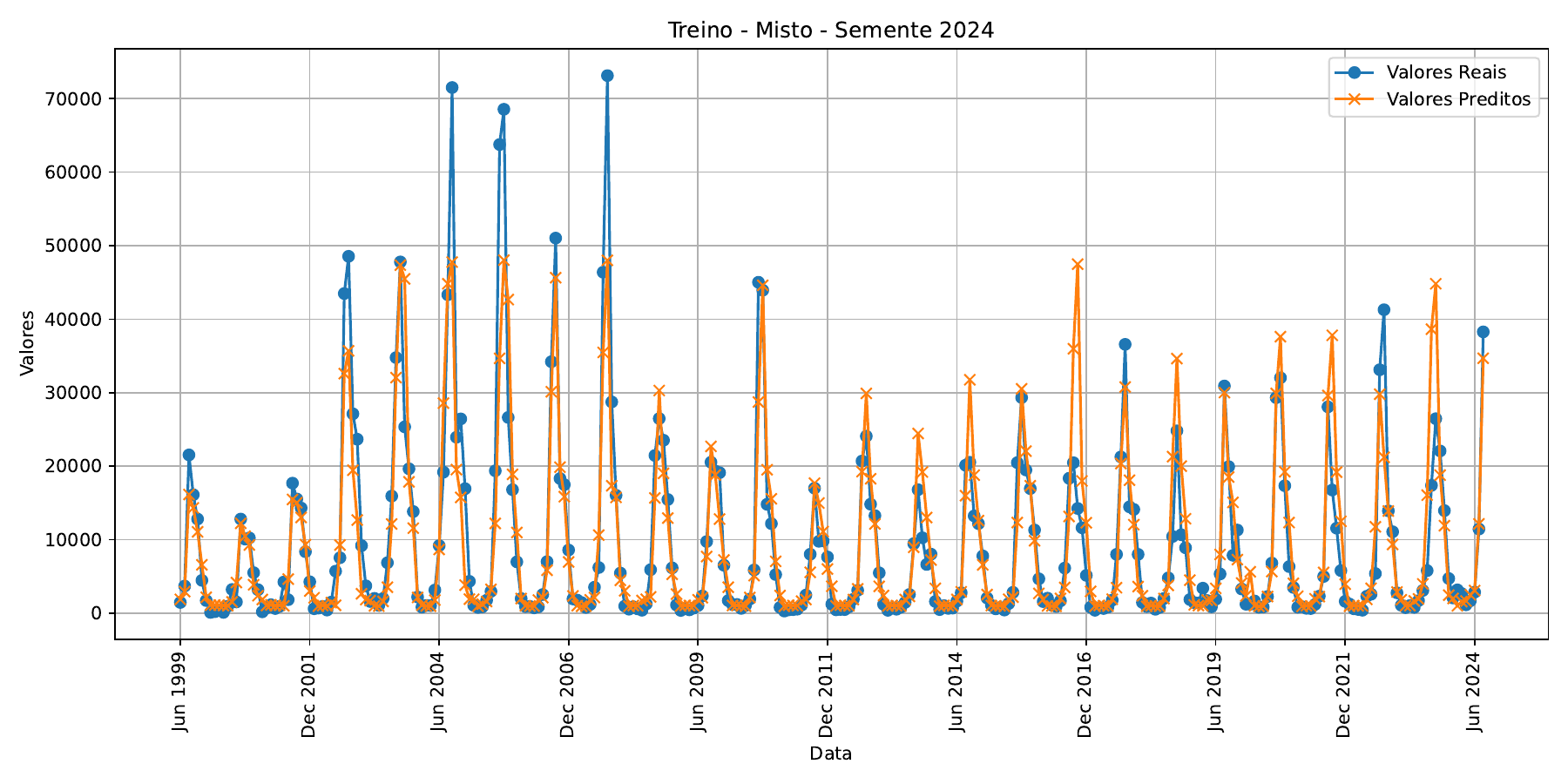}
  \caption{Figura ilustrativa do conjunto de treino em que é apresentados os dados reais da série e os preditos (Semente 2024)}
  \label{fig:sub10}
\end{subfigure}
\hfill
\begin{subfigure}{0.48\textwidth}
  \centering
  \includegraphics[width=\linewidth]{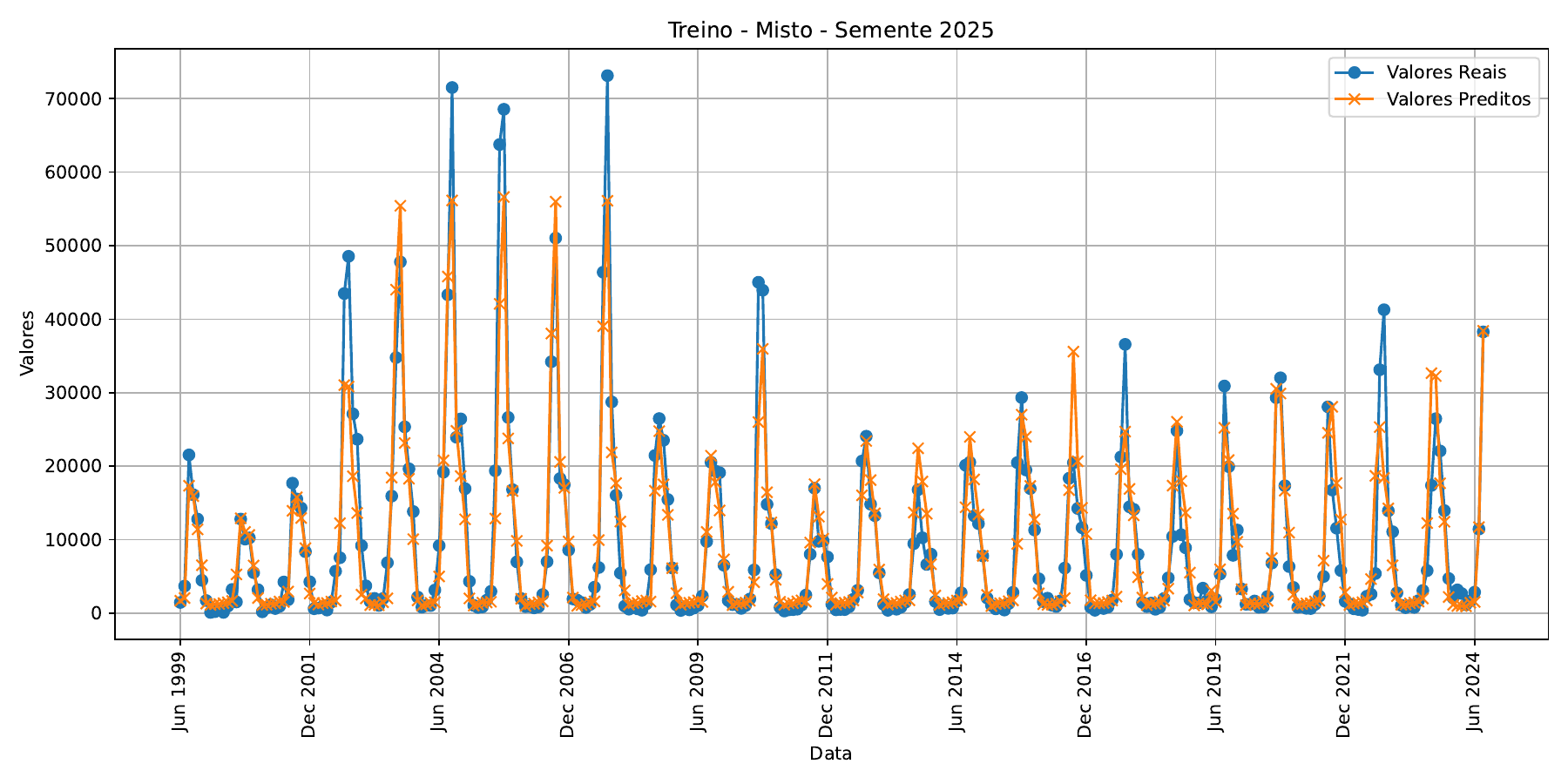}
  \caption{Figura ilustrativa do conjunto de treino em que é apresentados os dados reais da série e os preditos (Semente 2025)}
  \label{fig:sub20}
\end{subfigure}

\begin{subfigure}{0.48\textwidth}
  \centering
  \includegraphics[width=\linewidth]{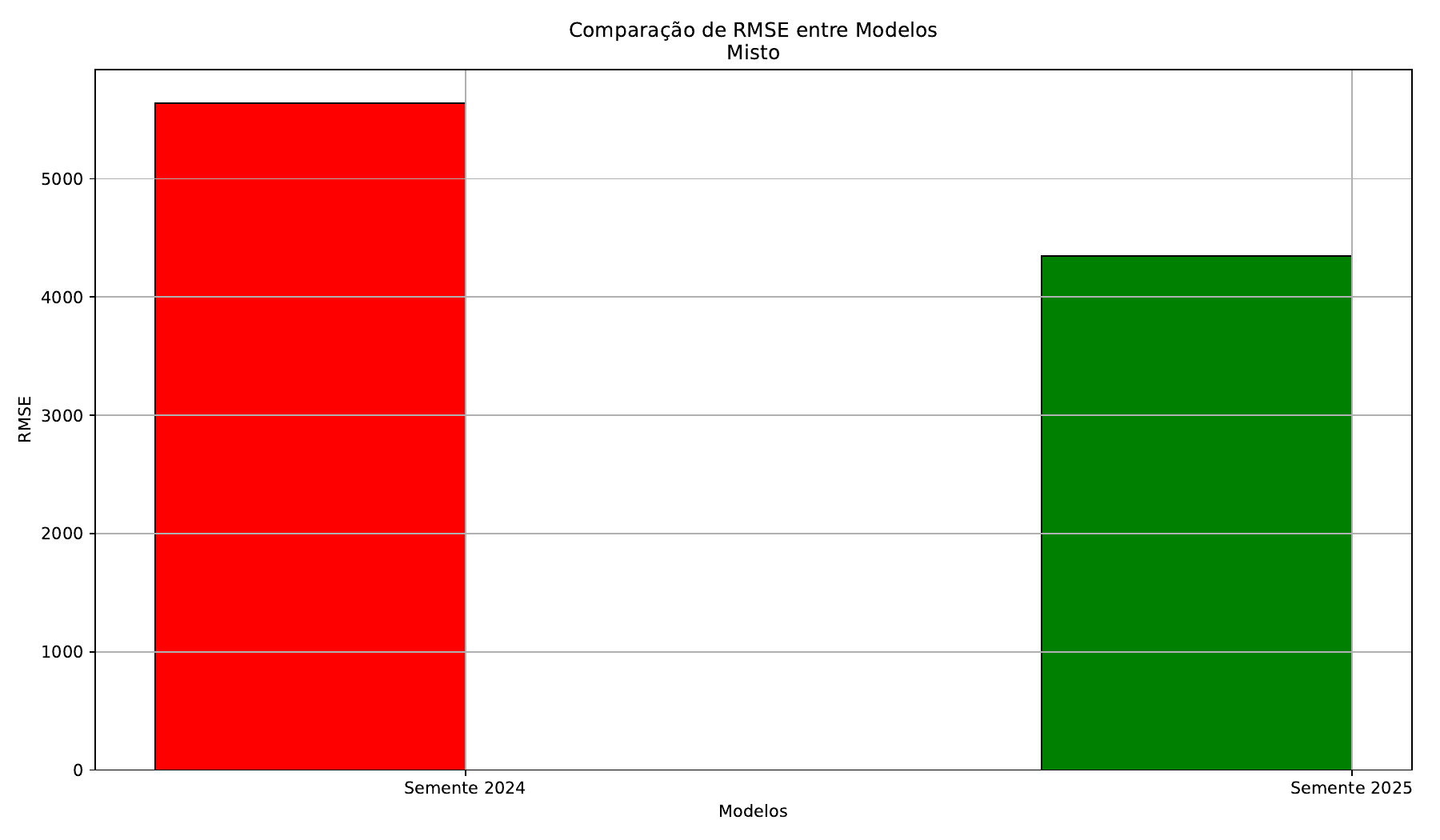}
  \caption{Figura ilustrativa da métrica raíz do erro quadrático médio (RMSE) - Tabela \ref{tab:dados_completos}}
  \label{fig:sub30}
\end{subfigure}
\hfill
\begin{subfigure}{0.48\textwidth}
  \centering
  \includegraphics[width=\linewidth]{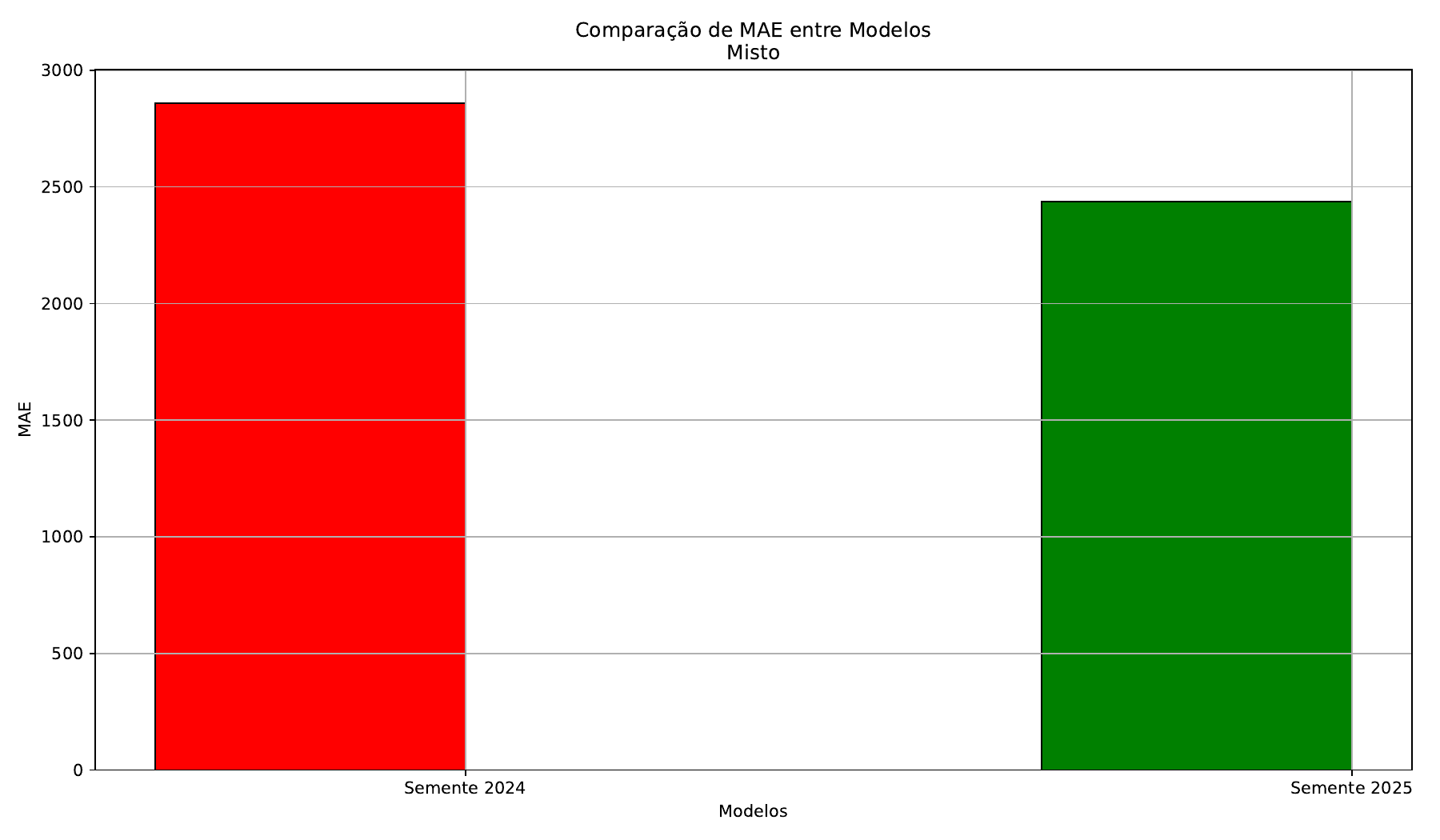}
  \caption{Figura ilustrativa da métrica erro absoluto médio (MAE) - Tabela \ref{tab:dados_completos}}
  \label{fig:sub40}
\end{subfigure}

\begin{subfigure}{0.48\textwidth}
  \centering
  \includegraphics[width=\linewidth]{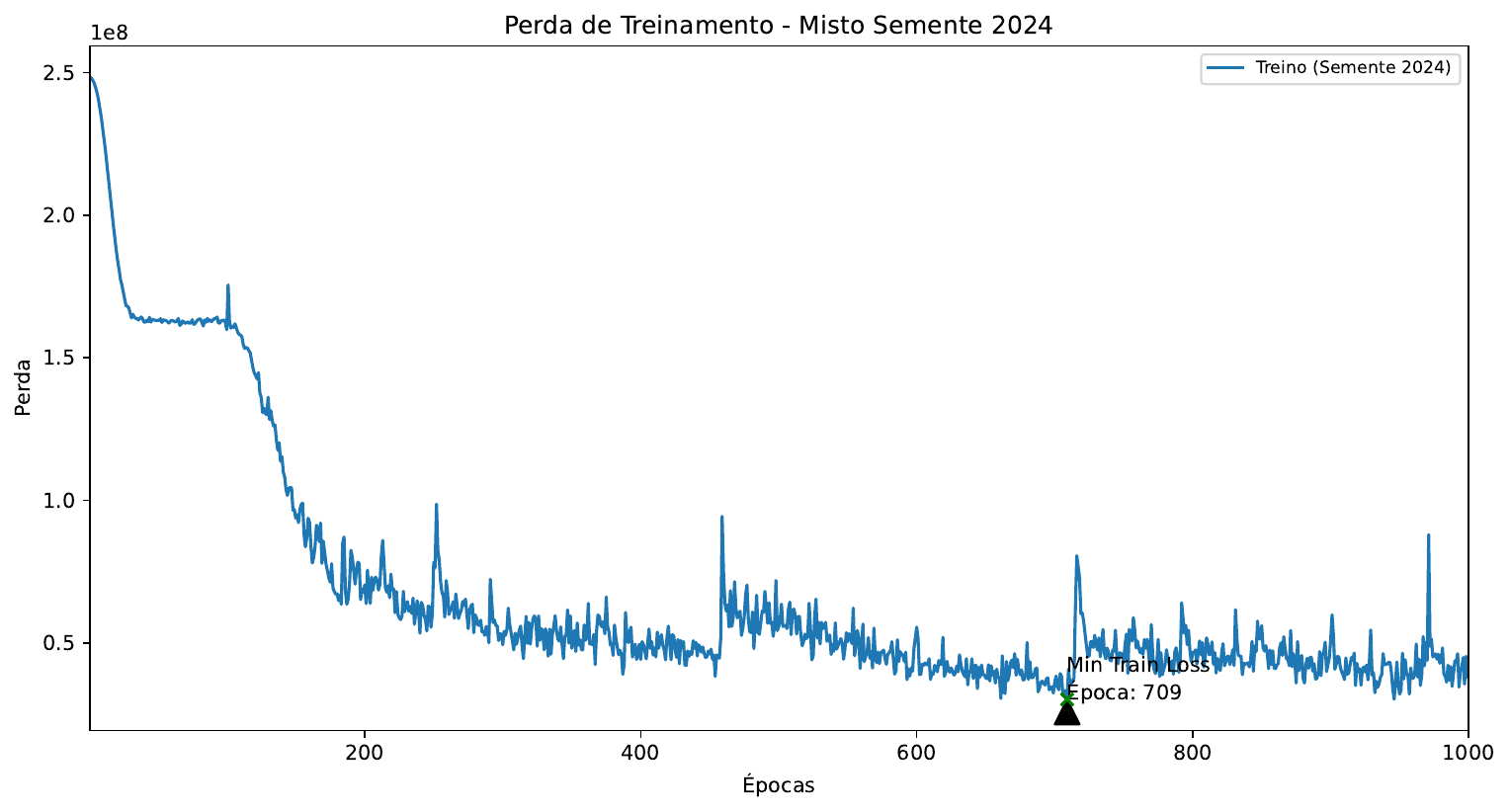}
  \caption{Figura ilustrativa da perda (\textit{Loss}) que foi realizado no conjunto de treino (Semente 2024)}
  \label{fig:sub50}
\end{subfigure}
\hfill
\begin{subfigure}{0.48\textwidth}
  \centering
  \includegraphics[width=\linewidth]{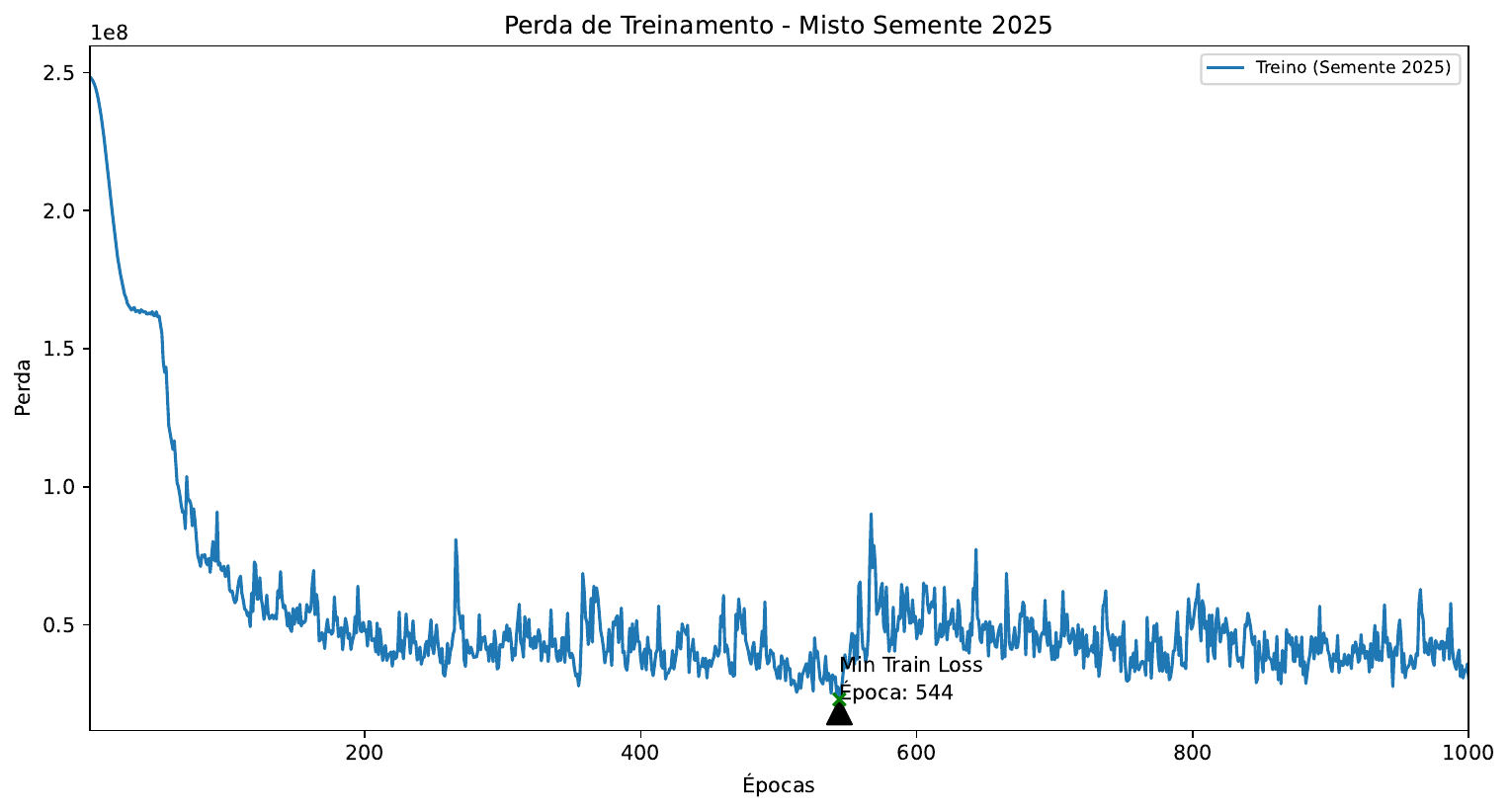}
  \caption{Figura ilustrativa da perda (\textit{Loss}) que foi realizado no conjunto de treino (Semente 2025)}
  \label{fig:sub60}
\end{subfigure}

\caption*{Fonte: Autoria própria} 
\label{fig:imagensTreino}
\end{figure}

\begin{table}[H]
	\centering
	\caption{Desempenho das métricas RMSE e MAE do modelo LSTM+GRU ilustradas nas Figuras - \ref{fig:sub30} e \ref{fig:sub40}}
	\begin{tabular}{ccccc}
		\hline
		Semente & RMSE Treino & MAE Treino \\
		\hline
		2024 & 5548 & 2893 \\
		2025 & 4137 & 2488 \\
		\hline
	\end{tabular}
	\label{tab:dados_completos}\\
	\noindent  Fonte: Autor, baseado nos resultados das métricas
\end{table}

A Figura \ref{fig:imagensTreino} apresenta o treinamento do modelo utilizando o conjunto completo de dados, abrangendo o período de junho de 1998 até agosto de 2024. Esta abordagem permite avaliar a performance do modelo em toda a série histórica, seguindo as melhores práticas para séries temporais em machine learning, onde é crucial treinar o modelo com todos os dados disponíveis para realizar previsões futuras. Os gráficos \ref{fig:sub10} e \ref{fig:sub20} mostram a comparação entre os dados reais e as previsões para o conjunto de treino, utilizando as sementes 2024 e 2025, respectivamente. Estes gráficos ilustram a capacidade do modelo em capturar a tendência geral dos dados ao longo do tempo. Os gráficos \ref{fig:sub30} e \ref{fig:sub40} apresentam as métricas de desempenho do modelo para as sementes 2024 e 2025, com a raiz do erro quadrático médio (RMSE) e o erro absoluto médio (MAE), respectivamente. Essas métricas são fundamentais para avaliar a precisão das previsões do modelo, com o RMSE fornecendo uma medida penalizada dos erros maiores e o MAE oferecendo uma visão geral das diferenças médias. Finalmente, os gráficos \ref{fig:sub50} e \ref{fig:sub60} ilustram a evolução da perda (\textit{Loss}) durante o treinamento para as sementes 2024 e 2025. A perda (\textit{Loss}) é uma métrica importante que representa o erro médio entre os valores reais e as previsões do modelo em cada época de treinamento. Esses gráficos mostram como a perda varia ao longo das épocas, refletindo o ajuste contínuo dos parâmetros do modelo para minimizar o erro. A combinação dessas análises permite uma visão abrangente do desempenho do modelo treinado com o conjunto completo de dados, confirmando a eficácia da abordagem adotada e a capacidade do modelo de generalizar bem para previsões futuras.

\subsection{Previsão}

\begin{figure}[H]
    \centering
    \caption{Figura ilustrativa do conjunto de treino com previsões de 12 \textit{lags} (meses) à frente, considerando que o modelo diz respeito à semente de valor 2025.}
    \begin{minipage}{0.48\linewidth}
        \centering
        \includegraphics[width=\linewidth]{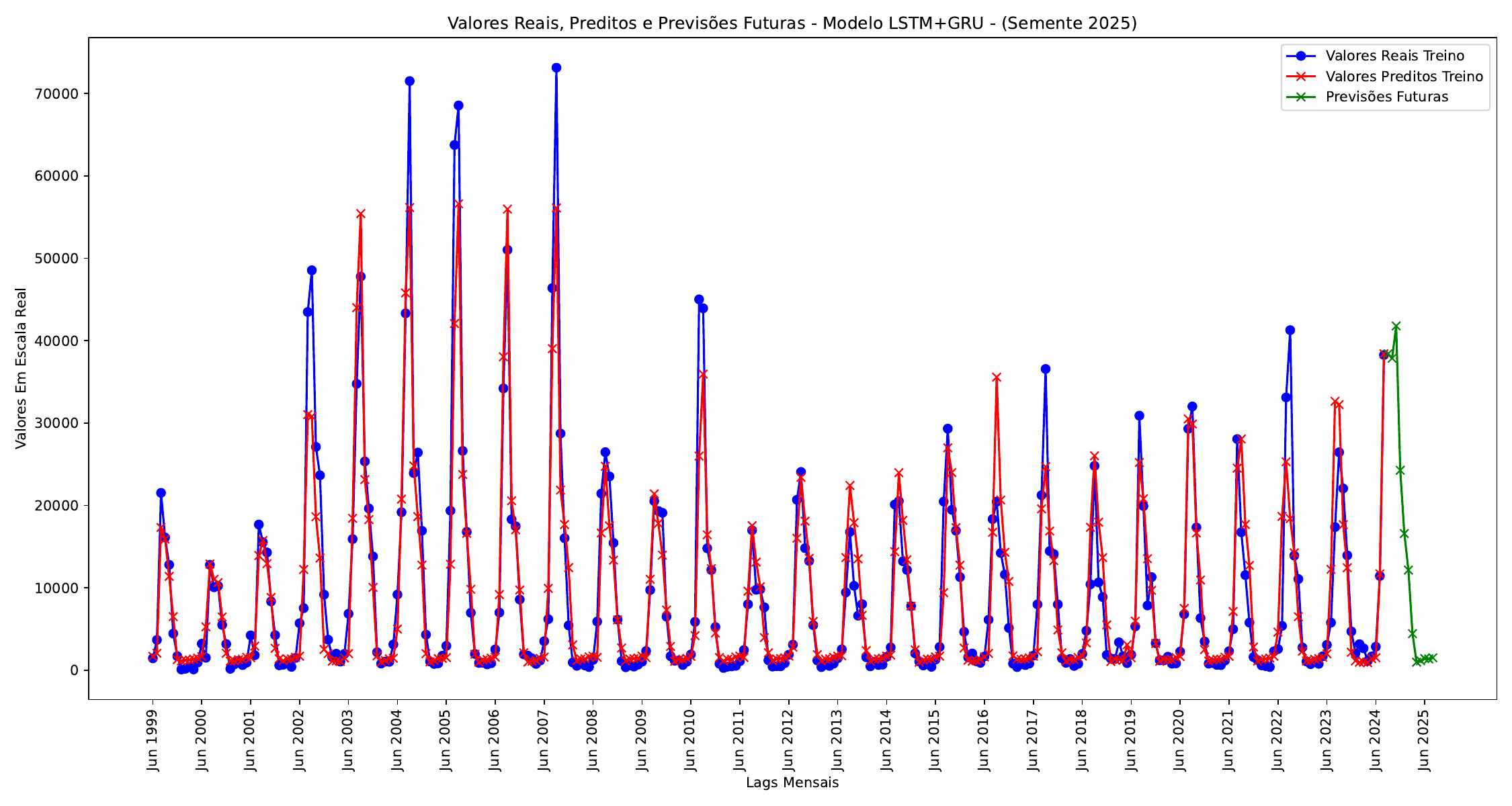}
        \caption*{Resultados de treinamento e previsões para a semente 2025.}
    \end{minipage}
    \hfill
    \begin{minipage}{0.48\linewidth}
        \centering
        \includegraphics[width=\linewidth]{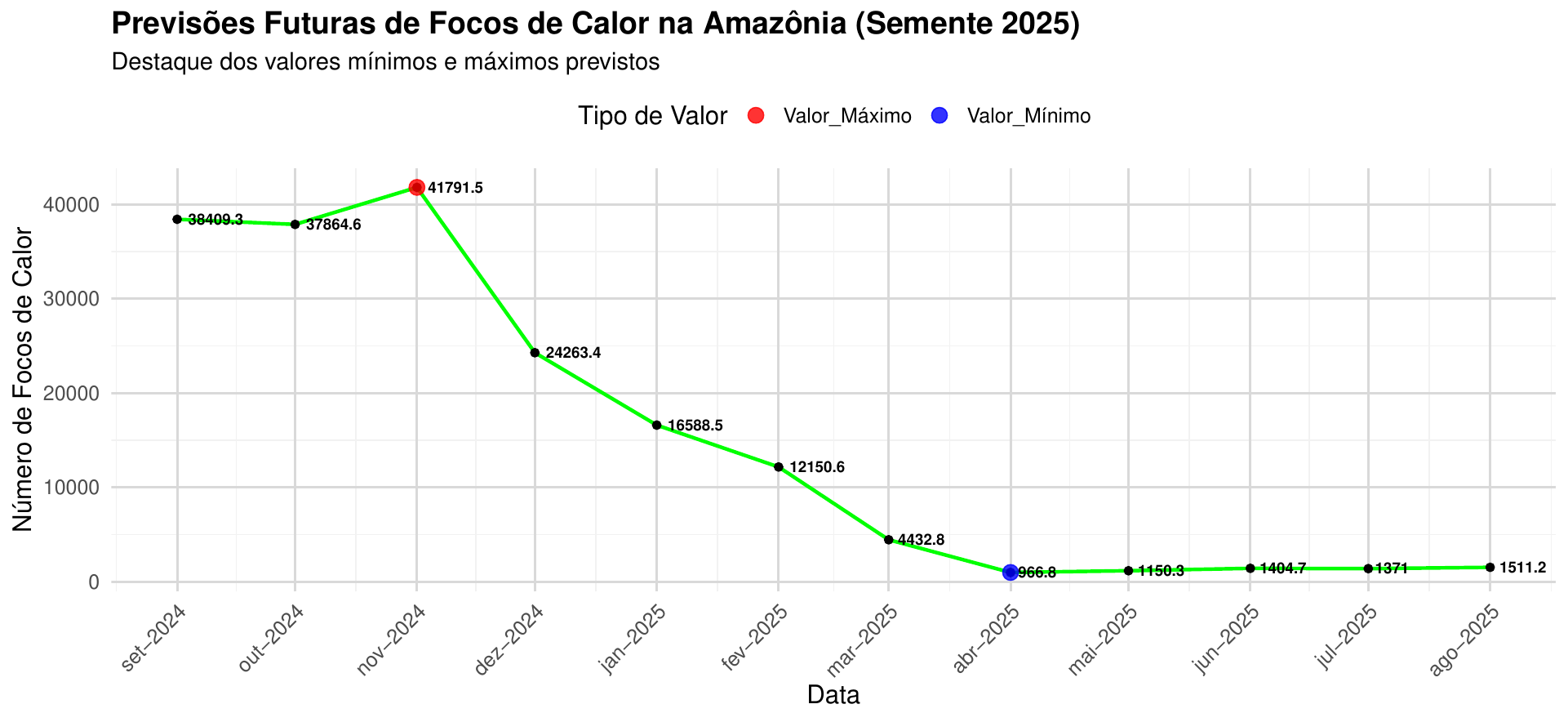}
        \caption*{Série temporal com previsões de 12 meses à frente para a semente 2025.}
    \end{minipage}
    \caption*{Fonte: Autoria própria} 
    \label{fig:treinoPrevisao}
\end{figure}

Então, a partir do modelo treinado e identificando o \textit{best model} — o ponto com o menor valor de métrica, como ilustrado nas Figuras \ref{fig:sub50} e \ref{fig:sub60} — podemos observar que esse ponto representa a configuração de parâmetros que resultou nos menores valores de erro absoluto médio (MAE) e erro quadrático médio (RMSE). Esse modelo otimizado foi utilizado para realizar previsões de 12 meses à frente. A Figura \ref{fig:treinoPrevisao} mostra a série temporal desde junho de 1999 até agosto de 2024, e as previsões geradas se estendem de setembro de 2024 até agosto de 2025. Para detalhes adicionais sobre o processo de previsão e a implementação, consulte a seção \ref{sec:previsoes} desse material.

\section{Conclusão}

Este estudo avaliou o desempenho de redes neurais recorrentes, combinando as arquiteturas \textit{Long Short-Term Memory} (LSTM) e \textit{Gated Recurrent Unit} (GRU), na previsão de focos ativos detectados pelo satélite \textit{AQUA\_M-T} na Amazônia. A análise demonstrou que essa combinação é eficaz na captura de padrões temporais complexos, com a escolha final do modelo baseada no Erro Absoluto Médio (MAE). As previsões geradas pelos modelos combinados apresentaram um desempenho elevado, especialmente em séries temporais com forte sazonalidade, como é o caso da série histórica dos focos ativos na Amazônia. Além disso, o estudo destacou a importância da configuração meticulosa dos modelos e do treinamento com validação cruzada para garantir boas práticas de modelagem. A simplificação da compreensão das redes neurais e do processo de aprendizado torna o tema mais acessível a alunos e pesquisadores sem experiência prévia no assunto. A análise descritiva identificou os meses de máxima e mínima incidência de focos ativos de 1998 a 2024, oferecendo percepções estratégicas sobre esses eventos. A semente escolhida para treinar o modelo completo e realizar previsões futuras, entre as sementes de número 2024 e 2025, foi a 2025. Esta semente apresentou um erro absoluto médio de aproximadamente 2.500 focos ativos. O modelo previu que o mês de novembro de 2024 será o maior pico registrado, com 41.791,5 mil focos ativos na região da Amazônia. Esta previsão indica uma anomalia em relação à série histórica completa, que tradicionalmente registra picos mais altos em agosto ou setembro. A previsão de novembro sugere uma mudança na tendência de pico de focos ativos, enquanto os menores registros permanecem consistentes com o padrão histórico, tipicamente observados no primeiro semestre. Foi observado que, ao modelar dados de contagem, a utilização de técnicas apropriadas e referências confiáveis permite estruturar uma base de dados com grande assimetria e treinar modelos que convergem para previsões que seguem a tendência geral da série temporal, independentemente da escala dos dados.

\printbibliography[title={Referências}]

@article{Schmidhuber1997,
	author = {Hochreiter, S. and Schmidhuber, J.},
	title = {Long Short-Term Memory},
	journal = {Neural Computation},
	volume = {9},
	number = {8},
	pages = {1735--1780},
	year = {1997},
	month = {November},
	issn = {0899-7667},
	doi = {10.1162/neco.1997.9.8.1735},
	url = {https://doi.org/10.1162/neco.1997.9.8.1735},
	eprint = {https://direct.mit.edu/neco/article-pdf/9/8/1735/813796/neco.1997.9.8.1735.pdf},
}

@article{Greff2017,
	author = {Greff, K. and Srivastava, R. K. and Koutník, J. and Steunebrink, B. R. and Schmidhuber, J.},
	title = {LSTM: A Search Space Odyssey},
	journal = {IEEE Transactions on Neural Networks and Learning Systems},
	volume = {28},
	number = {10},
	pages = {2222--2232},
	year = {2017},
	doi = {10.1109/TNNLS.2016.2582924},
	url = {https://doi.org/10.48550/arxiv.1503.04069},
	note = {Version: arXiv:1503.04069v2 [cs.NE]; Submitted on March 13, 2015 (v1), last revised on October 4, 2017 (this version, v2)},
}

@article{Graves2013,
	author = {Graves, A. and Mohamed, A. and Hinton, G.},
	title = {Speech Recognition with Deep Recurrent Neural Networks},
	journal = {arXiv preprint},
	volume = {abs/1303.5778},
	pages = {1--5},
	year = {2013},
	url = {https://doi.org/10.48550/arXiv.1303.5778},
	eprinttype = {arXiv},
	eprint = {1303.5778},
	note = {Submitted on March 22, 2013},
}

@article{Kingma2014,
	author = {Kingma, D. P. and Ba, J. L.},
	title = {Adam: A Method for Stochastic Optimization},
	journal = {arXiv preprint},
	archivePrefix = {arXiv},
	primaryClass = {cs.LG},
	pages = {1--15},
	year = {2014},
	url = {https://arxiv.org/abs/1412.6980}, 
	note = {Submitted on December 22, 2014 (v1), last revised on January 30, 2017 (v9)},
}

@book{Goodfellow2016,
	title = {Deep Learning},
	author = {Goodfellow, I. and Bengio, Y. and Courville, A.},
	publisher = {MIT Press},
	year = {2016},
	pages = {1--223},
	url = {https://arxiv.org/abs/1412.6980},
}

@article{Cheng2024,
	author = {Cheng, L. and Pandey, A. and Xu, B. and Delbruck, T. and Liu, S. C.},
	title = {Dynamic Gated Recurrent Neural Network for Compute-efficient Speech Enhancement},
	journal = {Proceedings of Interspeech 2024},
	year = {2024},
	pages = {1--5},
	doi = {10.21437/Interspeech.2024-958},
	url = {http://dx.doi.org/10.21437/Interspeech.2024-958},
}

@article{Ruder2016,
	title = {An Overview of Gradient Descent Optimization Algorithms},
	author = {Ruder, S.},
	year = {2017},
	journal = {arXiv preprint},
	pages = {1--14},
	eprint = {1609.04747},
	archivePrefix = {arXiv},
	primaryClass = {cs.LG},
	url = {https://arxiv.org/abs/1609.04747},
}

@article{JunyoungChung2014,
	author = {Chung, J. and Gulcehre, C. and Cho, K. and Bengio, Y.},
	title = {Empirical Evaluation of Gated Recurrent Neural Networks on Sequence Modeling},
	journal = {CoRR},
	volume = {abs/1412.3555},
	pages = {1--9},
	year = {2014},
	url = {http://arxiv.org/abs/1412.3555},
	eprinttype = {arXiv},
	eprint = {1412.3555},
	bibsource = {dblp computer science bibliography, https://dblp.org},
}

@misc{INPE2024,
	author = {INPE},
	title = {Situação Atual das Queimadas no Brasil},
	year = {2024},
	url = {http://terrabrasilis.dpi.inpe.br/queimadas/situacao-atual/estatisticas/estatisticas_estados/},
	note = {Acessado em 12/08/2024},
}

@misc{SEIA2024,
	author = {SEIA},
	year = {2024},
	title = {Situação Atual das Queimadas Florestais},
	url = {http://www.seia.ba.gov.br/monitoramento-ambiental/focos-de-calor},
	note = {Acessado em 04/09/2024},
}

@misc{Python2024,
	author = {Python},
	year = {2024},
	title = {Python Programming Language},
	url = {https://www.python.org/},
	note = {Acessado em 04/09/2024},
}

@misc{FrançoisChollet2024,
	author = {Keras},
	title = {Keras Documentation},
	year = {2024},
	url = {https://keras.io/},
}

@misc{KerasDevelopers2024,
	author = {KerasDevelopers},
	title = {Initializers in Keras},
	howpublished = {\url{https://keras.io/api/layers/initializers/}},
	year = {2024},
	note = {Acessado em: 2024-09-01}
}

@manual{RCoreTeam2024,
	title     = {R: A Language and Environment for Statistical Computing},
	author    = {R},
	year      = {2024},
	publisher = {R Foundation for Statistical Computing},
	url = {https://www.r-project.org/},
	note = {Acessado em 01/09/2024}
}

@manual{shiny2024,
	title     = {Shiny: Web Application Framework for R},
	author    = {ShinyApp},
	year      = {2024},
	version   = {1.7.4},
	publisher = {R Package},
	url       = {https://CRAN.R-project.org/package=shiny},
	note = {Acessado em 01/09/2024}
	}

@book{Geron2017,
	author = {Géron, A.},
	title = {Hands-On Machine Learning with Scikit-Learn, Keras, and TensorFlow},
	year = {2017},
	publisher = {O'Reilly Media, Inc.},
	address = {Sebastopol, CA},
	url = {https://www.oreilly.com/library/view/hands-on-machine-learning/9781492032632/},
	pages = {1-707} 
}

@book{ScikitLearn2011,
	author = {Pedregosa, F. and Varoquaux, G. and Gramfort, A. and Michel, V. and Thirion, B. and Grisel, O. and Blondel, M. and Prettenhofer, P. and Weiss, R. and Dubourg, V. and Vanderplas, J. and Passos, A. and Cournapeau, D. and Brucher, M. and Perrot, M. and Duchesnay, E.},
	title = {Scikit-learn: Machine Learning in Python},
	journal = {Journal of Machine Learning Research},
	volume = {12},
	pages = {2825-2830},
	year = {2011}
}

@book{BoxJenkins1976,
	author = {Box, G. E. P. and Jenkins, G. M.},
	title = {Time Series Analysis: Forecasting and Control},
	publisher = {Holden-Day},
	year = {1976},
	address = {San Francisco},
	pages = {1-575},
}

@book{JamesHamilton1994,
	author = {Hamilton, J. D.},
	title = {Time Series Analysis},
	publisher = {Princeton University Press},
	year = {1994},
	address = {Princeton},
	pages = {1-799},
}

@book{PeterBrockwell2002,
	author = {Brockwell, P. J. and Davis, R. A.},
	title = {Introduction to Time Series and Forecasting},
	publisher = {Springer},
	year = {2002},
	address = {New York},
	pages = {1-436},
}

\end{document}